\documentclass{article}

\usepackage[final]{neurips_2026}

\usepackage[utf8]{inputenc} % allow utf-8 input
\usepackage[T1]{fontenc}    % use 8-bit T1 fonts
\usepackage{url}            % simple URL typesetting
\usepackage{booktabs}       % professional-quality tables
\usepackage{amsfonts}       % blackboard math symbols
\usepackage{nicefrac}       % compact symbols for 1/2, etc.
\usepackage{microtype}      % microtypography
\usepackage{xcolor}         % colors

\definecolor{citeblue}{rgb}{0.21,0.49,0.74}
\usepackage[pagebackref=false,breaklinks,colorlinks,citecolor=citeblue,bookmarks=false]{hyperref}
\usepackage{graphicx}
\usepackage{svg}
\usepackage{multirow}
\usepackage{colortbl}
\usepackage{soul}
\usepackage{wrapfig}
\usepackage{adjustbox}
\usepackage[most]{tcolorbox}
\usepackage{textcomp}
\usepackage{subcaption}
\usepackage{algorithm}
\usepackage{algpseudocode}
\usepackage{tabularx}
\usepackage{makecell}
\usepackage{float}
\usepackage{placeins}

\definecolor{secondcolor}{RGB}{223,235,254}
\definecolor{bestcolor}{RGB}{254,226,226}
\definecolor{promptblue}{RGB}{2,45,103}
\definecolor{algorithmblue}{RGB}{88, 137, 204}

\definecolor{teachercolor}{RGB}{237,233,254}

\newcommand{\tch}[1]{{\setlength{\fboxsep}{1.2pt}\colorbox{teachercolor}{#1}}}

\usepackage{array}
\usepackage{xfp}

\definecolor{MainBlue}{RGB}{2,45,103}
\definecolor{MainRed}{RGB}{138,46,44}

\newcommand{\tokheat}[2]{%
  \begingroup
  \edef\shade{\fpeval{min(82, max(6, round(100 * abs(#2) / 1.8)))}}%
  \ifdim #2 pt < 0pt
    \colorbox{MainRed!\shade!white}{\strut\texttt{#1}}%
  \else
    \colorbox{MainBlue!\shade!white}{\strut\texttt{#1}}%
  \fi
  \endgroup
}

\newcommand{\tokcell}[2]{%
  \begingroup
  \edef\shade{\fpeval{min(85, max(3, round(100 * abs(#2) / 16.625)))}}%
  \ifdim #2 pt < 0pt
    \colorbox{MainRed!\shade!white}{\strut\texttt{#1}}%
  \else
    \colorbox{MainBlue!\shade!white}{\strut\texttt{#1}}%
  \fi
  \endgroup
}

\algrenewcommand\algorithmiccomment[1]{%
  \hfill\textcolor{algorithmblue}{\(\triangleright\) #1}%
}

\newcommand{\best}[1]{\cellcolor{bestcolor}\textbf{#1}}
\newcommand{\second}[1]{\cellcolor{secondcolor}\underline{#1}}
\newcommand{\method}{RetireOPD}

\title{\method{}: Self-Retiring On-Policy Distillation for Agentic Reinforcement Learning}

\author{%
  \textbf{Yan Yu}$^{1}$\thanks{Equal contribution.},
  \textbf{Zhengxi Lu}$^{1}$\footnotemark[1],
  \textbf{Yizhou Liu}$^{1}$,
  \textbf{Yichen Pan}$^{1}$,
  \textbf{Aozhe Wang}$^{1}$,
  \textbf{Qipeng Chen}$^{1}$,\\
  \textbf{Hua Yang}$^{2}$,
  \textbf{Wenqi Zhang}$^{1}$,
  \textbf{Qianglong Chen}$^{2}$,
  \textbf{Yongliang Shen}$^{1}$\thanks{Corresponding author.}\\
  $^{1}$Zhejiang University \qquad
  $^{2}$Alibaba Group\\
  \texttt{\{yuy2003, zhengxilu, syl\}@zju.edu.cn}
}

\makeatletter
\def\@noticestring{}
\makeatother

\begin{document}
\maketitle
\begin{abstract}
  Multi-turn agents trained with reinforcement learning (RL) receive a single scalar reward per trajectory, which motivates self on-policy distillation (OPD) to supply dense token-level supervision from a self-teacher with privileged task skills, letting a skill-free student internalize them.
This recipe, however, is undermined by two findings in agentic tasks: privileged information alone does not always make a teacher reliable, and the benefit of teacher supervision is stage-dependent.
We therefore propose \textbf{\method{}} (Self-\textbf{Retiring} \textbf{On-Policy Distillation}), which first optimizes a decoupled, skill-conditioned teacher with environment rewards and then trains a skill-free student jointly with RL and OPD.
Rather than following a predefined distillation schedule, \method{} adopts \textbf{Adaptive Retirement}: the student drops the teacher on its own once their discrepancy stops shrinking and it reaches a target fraction of the teacher's success rate, after which training proceeds with RL alone.
Across Qwen2.5 models from 1.5B to 7B, \method{} improves ALFWorld success rate over RL baseline by 14.1\% to 18.8\% and WebShop accuracy by 11.8\% to 19.0\%, and surpasses its own skill-conditioned teacher in every setting.
Code is available at \href{https://github.com/ZJU-REAL/SDAR}{https://github.com/ZJU-REAL/SDAR}.
\end{abstract}
\vspace{-0.7mm}
\begin{figure}[htbp]
    \centering
    \includegraphics[
        width=1\linewidth,
        trim=0 0cm 0 0,
        clip
    ]{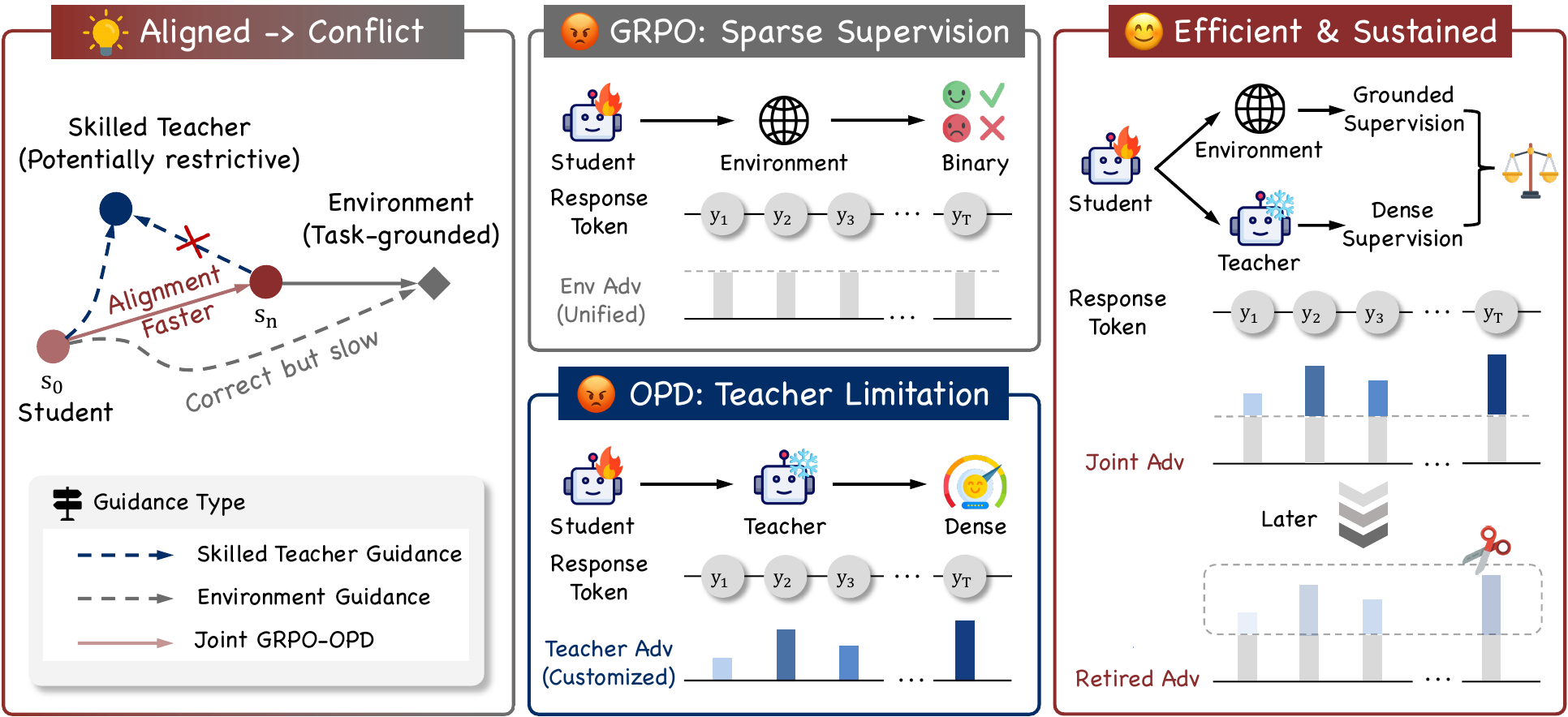}
    \caption{\textbf{Left:} GRPO and teacher guidance align early but later diverge, making teacher matching restrictive. \textbf{Middle:} GRPO's supervision is sparse and task-level, while OPD is dense but teacher-bound. \textbf{Right:} \method{} uses both early and retires the teacher at the conflict point, surpassing it.}
    \label{fig:motivation}
\end{figure}
\section{Introduction}

Reinforcement learning with verifiable rewards (RLVR) is the standard approach for training large language model (LLM) agents on multi-turn tasks~\citep{singh2025openai,shao2024deepseekmath}, where a single outcome reward per trajectory leaves the intermediate decisions of a long interaction unsupervised. On-Policy Distillation (OPD) complements this sparse reward with dense token-level supervision from a teacher on trajectories sampled by the student~\citep{zeng2026glm,team2026qwen3,xu2026deepseek,lu2025opd}. In agentic training, the teacher is commonly the same model conditioned on privileged context that is available only during training, such as retrieved task skills~\citep{zhao2026self,lu2026self}. The student is trained to reproduce the teacher's behavior without access to this context, so that the skills are internalized into its parameters and no additional context is required at inference~\citep{lu2026skill0,wang2026skill}.

\begin{figure}[h]

\begin{center}
\includegraphics[width=1\linewidth]{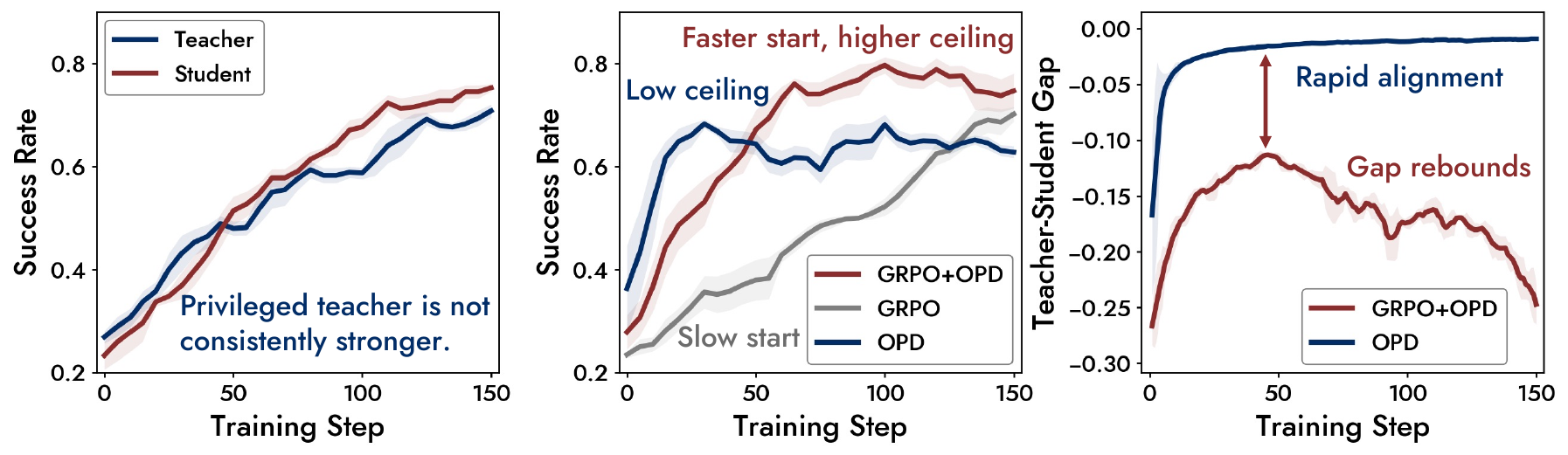}
\end{center}

\caption{Training dynamics of Qwen2.5-3B-Instruct on ALFWorld. \textbf{Left}: Success rates of the privileged teacher and student branches using the same model. \textbf{Middle}: Success rate under GRPO, OPD and GRPO+OPD. \textbf{Right}: Teacher-student discrepancy under OPD and GRPO+OPD.}
\label{fig:comparison}

\end{figure}

This paradigm rests on two assumptions. The teacher is assumed to be reliably better than the student because it sees the privileged context, and matching the teacher is assumed to remain useful for as long as training lasts~\citep{zhang2026verify,zhou2026sage}. They amount to two questions that current methods answer by assumption~\citep{lu2026self} or by a fixed schedule~\citep{tan2026atod,ding2026saf}: \emph{which teacher} the student should learn from, and \emph{for how long}. We examine these assumptions on ALFWorld and find that neither assumption holds (Figure~\ref{fig:comparison}).

\textbf{Privileged context alone does not make a teacher reliable.} Under joint Group Relative Policy Optimization (GRPO)~\citep{shao2024deepseekmath} and On-Policy Self-Distillation (OPSD)~\citep{zhao2026self}, where teacher and student are two branches of one shared policy~\citep{lu2026self}, the skill-conditioned branch does not consistently outperform the student it supervises (Figure~\ref{fig:comparison}, left). This suggests that the teacher does not effectively utilize the provided skill: although it receives privileged skill context, it has not been adequately trained to incorporate this information into its decisions. Simply increasing the model size does not resolve this issue either---an unoptimized 7B model prompted with the same skills reaches only 23.4\% success on ALFWorld. A teacher must therefore be \emph{optimized to use} its privileged context before it can provide effective supervision.

\textbf{The benefit of teacher supervision is stage-dependent.} Adding OPD to GRPO removes the slow start of pure GRPO and the low ceiling of pure OPD (Figure~\ref{fig:comparison}, middle). The teacher discrepancy first narrows and then widens (Figure~\ref{fig:comparison}, right). Once the student has internalized the behavior that the skills induce, reward optimization favors actions beyond the teacher, and the two gradients begin to conflict (Figure~\ref{fig:motivation}, left). Continuing to match the teacher beyond this point holds the student near the teacher's performance ceiling (Section~\ref{sec:dynamics}). A first-order analysis (Appendix~\ref{sec:theory}) shows that stagnating discrepancy implies diminishing OPD benefits or increasing gradient conflict. Teacher supervision is therefore \emph{temporary scaffolding}, and the question is when to remove it.

Existing work addresses this transition with predefined schedules, either annealing distillation~\citep{tan2026atod,ding2026saf} or switching from distillation to RL at a fixed step~\citep{ye2026opdsearch+,li2026sequential}.
Online decisions exist only at finer granularity, per prompt or per token~\citep{ding2026hdpo,lu2026self}, and none of them removes the teacher. In our experiments, the point at which the discrepancy stops decreasing varies from step 50 to step 90 across models and tasks (Appendix~\ref{sec:training_dynamics}), so a single schedule may withdraw guidance too early in some settings and too late in others. These observations suggest a simple principle: \emph{whether the teacher is still useful can be read from the training signal itself.} The discrepancy stops decreasing when the two objectives conflict, and the student's success rate relative to the teacher indicates whether the skills have been internalized. The transition should therefore be determined online rather than fixed in advance.

Motivated by these findings, we propose \textbf{\method{}} (Self-Retiring On-Policy Distillation), which internalizes privileged skills into a skill-free student and retires the teacher once its supervision no longer benefits the student. \method{} first decouples the teacher from the student and trains it with environment rewards under the skill context, so that distillation starts from a policy that has already learned to exploit the skills. The student is then trained jointly with GRPO and OPD, while the two signals identified above are monitored at fixed intervals. Once the discrepancy stops decreasing and the student's success rate reaches a set fraction of the teacher's, the teacher is retired and training proceeds with GRPO alone. The student is thus no longer constrained by the teacher, and no further teacher forward passes are required.

Across Qwen2.5-1.5B, 3B and 7B on ALFWorld~\citep{shridhar2020alfworld} and WebShop~\citep{yao2022webshop}, \method{} outperforms RL, distillation and hybrid baselines, improving ALFWorld success rate over GRPO by 14.1\% to 18.8\% and WebShop accuracy by 11.8\% to 19.0\%. It also surpasses its own skill-conditioned teacher in every setting. Our contributions are as follows:

\begin{itemize}
\item We identify two failure modes of privileged-information distillation for agents: an unoptimized skill-conditioned teacher is unreliable, and teacher matching conflicts with reward optimization once the student has internalized the teacher's knowledge.
\item We propose \method{}, which trains a same-capacity skill-conditioned teacher with environment rewards and retires it online once the teacher-student discrepancy stops decreasing and the student reaches a set fraction of the teacher's success rate.
\item Experiments across three model scales on ALFWorld and WebShop show that \method{} consistently outperforms RL, distillation and hybrid baselines while surpassing its own teacher, which validates its effectiveness on agentic tasks.
\end{itemize}
\section{Related Work}

\subsection{On-Policy Distillation with Privileged Teachers}
On-policy distillation (OPD) trains a student on its own samples with token-level feedback from a teacher~\citep{agarwal2024gkd,lu2025opd,song2026survey} and is now a standard post-training stage~\citep{xiao2026mimo,zeng2026glm,team2026kimi}. Without a stronger teacher, on-policy self-distillation (OPSD) conditions the same model on privileged information, such as a reference solution or textual feedback, and distills the conditioned branch into the unconditioned one~\citep{zhao2026self,hubotter2026reinforcement,ye2026policy,ma2026softpromptopd}. For agents, this privileged information is often a retrieved or hindsight skill~\citep{lu2026self,wang2026skill}, with teacher supervision controlled through token-level gating or shaping~\citep{lu2026self,zhang2026stepopsd}. Teacher reliability itself has received less attention, although guidance quality depends on compatibility with the student and on capability beyond it~\citep{li2026rethinking,liu2026privileged,chen2026look}, and only some tokens carry useful signal~\citep{xu2026tip,zhang2026unifyopd}. \method{} instead trains the teacher separately with environment rewards before distillation.

\subsection{Combining On-Policy Distillation with Reinforcement Learning}
RLVR is the dominant approach for multi-turn agents~\citep{shao2024deepseekmath,feng2026group,dong2026agentic,wang2025ragen,lu2025uis1}, but its trajectory-level reward is sparse, so hybrid methods add a distillation term or a teacher-shaped advantage to the RL objective~\citep{lu2026self,zhang2026learning,yang2026reconciling}. Because the two signals interfere~\citep{li2026sequential,pan2026rlcsd} as the student improves, several methods reduce the teacher's influence over training: annealing the distillation weight~\citep{tan2026atod,ding2026saf}, reducing the trajectory horizon exposed to the student~\citep{wang2026tcod,li2026guidedopd,xing2026trust}, or running distillation and RL in sequence~\citep{ye2026opdsearch+,kim2026opsd,dong2026rlstarts}. 
In all of these the transition is fixed before training. Signal-dependent decisions exist only at finer granularity, such as distilling on prompts where all rollouts fail~\citep{ding2026hdpo,han2026distill} or gating the teacher per token~\citep{lu2026self,wang2026teach}, and none removes the teacher. \method{} decides the global transition online from the teacher-student discrepancy and the student's relative competence.

\section{Method}
\label{sec:method}

\begin{figure}[t]

\begin{center}
\includegraphics[width=1\linewidth]{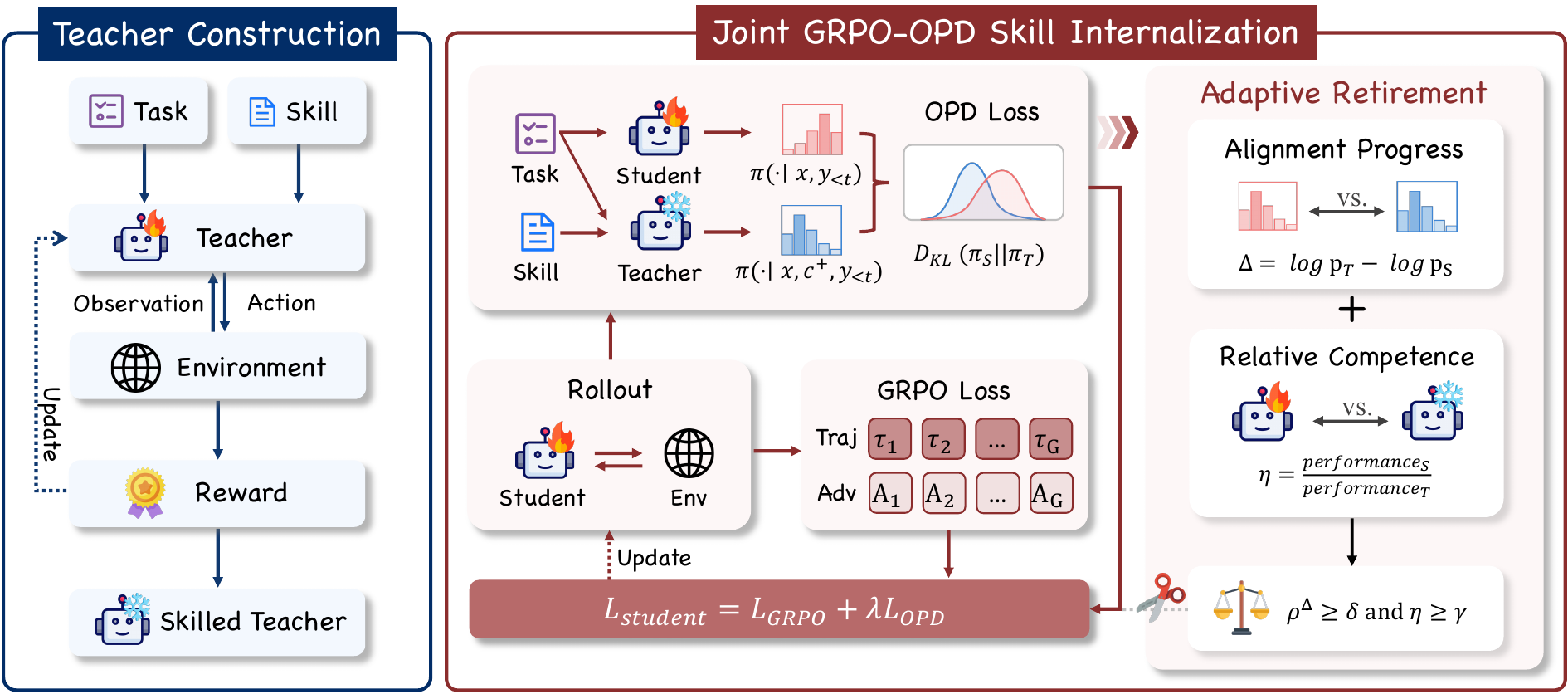}
\end{center}

\caption{\textbf{Overview of \method{}.} \textbf{(1) Teacher Construction}: We optimize a skill-conditioned teacher with environmental rewards. \textbf{(2) Joint Skill Internalization}: the student learns from both GRPO and OPD training. \textbf{(3) Adaptive Retirement}: OPD supervision is removed by current policy.}
\label{fig:method}

\end{figure}

As shown in Figure~\ref{fig:method}, our method consists of three stages:
\textbf{(1) Teacher construction}, optimizing a skill-conditioned teacher with environment rewards (Section~\ref{sec:teacher_construction});
\textbf{(2) Joint GRPO-OPD training}, where the skilled teacher supervises a skill-free student on student-generated trajectories (Section~\ref{sec:joint_grpo_opd}) and
\textbf{(3) Adaptive teacher retirement}, removing OPD once behavioral transfer stagnates and the student reaches sufficient task competence (Section~\ref{sec:exit}).

\paragraph{Task Definition.} We consider a multi-turn agent interacting with an environment over a sequence of decision steps. Given a task input $x \sim \mathcal{D}$, the agent iteratively generates responses based on the interaction history and receives environment feedback, forming a trajectory $\tau=(x,y)$, where $y=(y_1,\ldots,y_T)$ denotes the flattened sequence of all valid response tokens across interaction turns, and $R(\tau)$ is the scalar environment reward.

\subsection{Privileged Teacher Construction}
\label{sec:teacher_construction}

The base model is insufficient to utilize privileged skill context, necessitating teacher training. We therefore train a teacher policy to learn how to leverage skills for decision-making. Specifically, the teacher $\pi_\phi$ is provided with the privileged skill context $c^{+}$ during training. We optimize the teacher with GRPO~\citep{shao2024deepseekmath} to obtain a \textit{skilled teacher}, with the following objective:

\begin{equation}
\phi^{*}
=
\arg\max_{\phi}
\mathbb{E}_{x\sim\mathcal{D},\,\tau\sim\pi_{\phi}(\cdot\mid x,c^{+})}
\left[R(\tau)\right].
\end{equation}

Environment reward optimization encourages the teacher to turn privileged information into effective task behavior. After training, we freeze $\phi^{*}$ and denote the resulting policy as the skilled teacher $\pi_T$. The frozen teacher is subsequently used only to provide dense supervision for the student. Further details on teacher construction are provided in Appendix~\ref{app:teacher_construction}.

\subsection{Joint GRPO-OPD Optimization}
\label{sec:joint_grpo_opd}

With the skilled teacher obtained, we next train the student policy $\pi_\theta$ to internalize the teacher's behaviors. Following SDAR~\citep{lu2026self}, the student is trained without access to the privileged skill context $c^{+}$. For each input $x$, the student $\pi_\theta$ samples a group of $G$ trajectories. Based on the environment rewards, we first optimize the student with GRPO, which is defined as:

\begin{align}
\mathcal{L}_{\mathrm{GRPO}}(\theta)
={}&
-
\mathbb{E}_{x\sim\mathcal{D}}
\left[
\frac{1}{G}
\sum_{i=1}^{G}
\frac{1}{|y^{(i)}|}
\sum_{t=1}^{|y^{(i)}|}
\min
\left(
r_{t}^{(i)}(\theta)\hat{A}^{(i)},\,
\operatorname{clip}
\left(
r_{t}^{(i)}(\theta),1-\epsilon,1+\epsilon
\right)
\hat{A}^{(i)}
\right)
\right]
\nonumber\\
&+
\alpha_{KL}\,
\mathbb{E}_{x\sim\mathcal{D}}
\left[
\frac{1}{G}
\sum_{i=1}^{G}
\frac{1}{|y^{(i)}|}
\sum_{t=1}^{|y^{(i)}|}
D_{\mathrm{KL}}
\left(
\pi_\theta(\cdot\mid x,y_{<t}^{(i)})
\,\Vert\,
\pi_{\mathrm{ref}}(\cdot\mid x,y_{<t}^{(i)})
\right)
\right],
\end{align}

where $r_t^{(i)}(\theta)=\pi_\theta(y_t^{(i)}\mid x,y_{<t}^{(i)})/\pi_{\theta_{\mathrm{old}}}(y_t^{(i)}\mid x,y_{<t}^{(i)})$
is the token-level importance ratio.

For each student-generated trajectory, we compare the student and teacher distributions conditioned on the same trajectory prefix $y_{<t}$. The teacher provides supervision directly on states visited by the student. We define the OPD objective using the reverse KL divergence:

\begin{equation}
\mathcal{L}_{\mathrm{OPD}}(\theta)
=
\mathbb{E}_{x\sim\mathcal{D},\,y\sim\pi_\theta(\cdot\mid x)}
\left[
\frac{1}{|y|}
\sum_{t=1}^{|y|}
D_{\mathrm{KL}}
\left(
\pi_\theta(\cdot\mid x,y_{<t})
\,\Vert\,
\pi_T(\cdot\mid x,c^{+},y_{<t})
\right)
\right].
\end{equation}

To reduce the overhead of exact KL computation, we use a sampled-token approximation on student-generated trajectories, evaluating each $y_t$ under both the student and teacher distributions:

\begin{equation}
\Delta_t
=
\log \pi_T(y_t\mid x,c^{+},y_{<t})
-
\log \pi_\theta(y_t\mid x,y_{<t}).
\end{equation}

Finally, the full optimization objective is formulated as:
\begin{equation}\label{distillation target}
\mathcal{L}_{\mathrm{student}}(\theta)
=
\mathcal{L}_{\mathrm{GRPO}}(\theta)
+
\lambda \mathcal{L}_{\mathrm{OPD}}(\theta),
\end{equation}

where $\lambda$ controls the contribution of teacher supervision.

\subsection{Adaptive Teacher Retirement}
\label{sec:exit}
% Teacher supervision helps early but may later constrain reward-driven optimization. Since this transition depends on student learning dynamics, a fixed retiring step may be premature or delayed. We therefore determine the retiring point adaptively from training dynamics.

Teacher supervision is beneficial during the early stages of training, but can eventually constrain further performance gains by imposing a capability ceiling on the student. We therefore introduce an adaptive retirement mechanism that monitors training dynamics and determines when the teacher should be removed from the optimization process.

% We divide training into monitoring windows of $W$ steps. At step $n$, let $y_n^{(i)}$ be the $i$-th sampled trajectory. Using the token-level Teacher-Student log-probability gap $\Delta_t$ defined in Section~\ref{sec:joint_grpo_opd}, we write its $t$-th token gap as $\Delta_t^{(i)}$ and compute the step-level and window-level \textit{alignment progress} as

We partition training into monitoring windows of $W$ steps indexed by $m$. At step $n$, let $y_n^{(i)}$ denote the $i$-th sampled trajectory. Given the teacher-student log-probability gaps $\Delta_t^{(i)}$, we compute the step-level and window-level \textit{alignment progress} $K_n$ and $\bar{K}_m$:

% We partition training into monitoring windows of $W$ steps. Let $m$ denote the window index, and at step $n$, denote the $i$-th sampled trajectory by $y_n^{(i)}$. Given the Teacher-Student log-probability gaps $\Delta_t^{(i)}$ for the sampled trajectories, we then compute the step-level and window-level \textit{alignment progress} $K_n$ and $\bar{K}_m$:
\begin{equation}
    K_n=\frac{1}{G}\sum_{i=1}^{G}\frac{1}{|y_n^{(i)}|}\sum_{t=1}^{|y_n^{(i)}|}\Delta_t^{(i)},
\quad
\bar{K}_m
=
\frac{1}{W}
\sum_{n=mW}^{mW+W-1}
K_n.
\end{equation}

To capture when teacher-student alignment begins to stagnate or reverse, we define the \textit{rebound indicator} $\rho_m^{(K)}$ and \textit{relative competence} $\eta_m$ as

\begin{equation}
\label{eq:gap_change}
\rho_m^{(K)}
=
\frac{\bar{K}_m-\bar{K}_{m-1}}{\bar{K}_m},
\quad
\eta_m
=
\frac{SR_m+SR_{m-1}}{2SR_T},
\end{equation}

% where $SR_m$ and $SR_T$ denote the student and teacher success rates, respectively. A negative $\rho_m^{(K)}$ indicates a shrinking Teacher-Student gap, while a non-negative value indicates stagnation or rebound. $\eta_m$ measures the student's competence relative to the teacher. Teacher supervision is retired at the first monitoring window satisfying:

where $SR_m$ and $SR_T$ denote the student and teacher success rates, respectively. Teacher supervision is retired at the first monitoring window satisfying:

\begin{equation}\label{retire condition}
m^{*}
=
\inf
\left\{
m\ge 2:
\rho_m^{(K)}\ge\delta
\ \wedge\
\eta_m\ge\gamma
\right\}.
\end{equation}

% Here, $\delta$ and $\gamma$ are the discrepancy change and relative competence thresholds. Requiring both avoids premature retirement from transient fluctuations when the student is still weak. Once satisfied, OPD is removed and training continues with GRPO alone.

Here, $\delta$ and $\gamma$ denote the rebound indicator and relative competence thresholds, respectively. Requiring both criteria prevents premature retirement due to transient fluctuations, ensuring that the student has sufficiently benefited from teacher supervision before continuing independently.

\section{Experiments}
\subsection{Experiment Setup}
\paragraph{Benchmarks.}
We evaluate our method on two widely used interactive agent benchmarks: ALFWorld~\citep{shridhar2020alfworld} and WebShop~\citep{yao2022webshop}. 
ALFWorld is a text-based embodied environment where agents follow natural language instructions to complete multi-step household tasks, testing long-horizon planning, state tracking, and interaction with the environment.
WebShop simulates realistic online shopping scenarios, requiring agents to search, navigate, and select products that satisfy user-specified constraints, thereby evaluating goal-directed decision making and multi-step information gathering. Together, the two benchmarks cover complementary forms of interactive reasoning across embodied and web-based agentic tasks.

\paragraph{Baselines.}
We compare \method{} with four groups of baselines using Qwen2.5-1.5B / 3B / 7B-Instruct~\citep{qwen2.5}:
\textbf{(1) Prompting.} Vanilla directly evaluates the instruction-tuned model, while Skill-Prompt additionally provides task-relevant skills at inference time.
\textbf{(2) RL.} GRPO~\citep{shao2024deepseekmath}, GiGPO~\citep{feng2026group}, PPO~\citep{schulman2017proximal}, and RLOO~\citep{ahmadian2024back} are standard RL baselines, while Skill-GRPO~\citep{xia2026skillrl} performs GRPO with privileged skills and serves as our skill-conditioned teacher.
\textbf{(3) Distillation.} OPD~\citep{ye2026policy} and OPSD~\citep{zhao2026self} use dense teacher supervision.
\textbf{(4) Hybrid.} GRPO+OPD and GRPO+OPSD~\citep{lu2026self} combine RL with dense teacher supervision, while \method{} further removes teacher supervision adaptively.

\paragraph{Implementation.} We apply identically initialized teacher and student models at each scale. We set $W=5$, $\gamma=0.9$, and $\delta=0$. Following SDAR~\citep{lu2026self}, we set $\lambda=0.01$ and use its Keyword Matching strategy to retrieve privileged skills from the SkillBank of SkillRL~\citep{xia2026skillrl} in both environments. Additional details are provided in Appendix~\ref{app:detail}.

\begin{table*}[t]
\caption{Main results on ALFWorld and WebShop.
$^\ast$ means validation with skills.
The \tch{highlighted} suffix of a Distillation method denotes its teacher.
\sethlcolor{bestcolor}\hl{\textbf{Best}} and
\sethlcolor{secondcolor}\hl{\mbox{\underline{second-best}}}
results are highlighted.}
\label{tab:main-results}
\begin{center}
\fontsize{8}{9.5}\selectfont
\setlength{\tabcolsep}{4.2pt}
\renewcommand{\arraystretch}{1}
\resizebox{\textwidth}{!}{%
\begin{tabular}{ll*{9}{c}}
\toprule
& &
\multicolumn{7}{c}{\textbf{ALFWorld}} &
\multicolumn{2}{c}{\textbf{WebShop}} \\
\cmidrule(lr){3-9}
\cmidrule(lr){10-11}

\textbf{Method} &
\textbf{Type} &
\textbf{Pick} &
\textbf{Look} &
\textbf{Clean} &
\textbf{Heat} &
\textbf{Cool} &
\textbf{Pick2} &
\textbf{Avg} &
\textbf{Score} &
\textbf{Acc} \\
\midrule

\rowcolor{gray!15}
\multicolumn{11}{l}{\itshape Qwen2.5-1.5B-Instruct} \\

Vanilla~\citep{qwen2.5}
& Prompting
& 11.1 & 0.0 & 6.2 & 0.0 & 0.0 & 4.2 & 5.5 & 17.8 & 5.5 \\

Skill-Prompt$^\ast$~\citep{xia2026skillrl}
& Prompting
& 3.4 & 16.7 & 12.9 & 5.3 & 0.0 & 5.0 & 6.2 & 20.8 & 1.6 \\

GRPO~\citep{shao2024deepseekmath}
& RL
& 85.3 & 53.7 & 84.5 & 78.2 & 58.7 & 53.5 & 72.8 & 75.8 & 56.8 \\

GiGPO~\citep{feng2026group}
& RL
& 94.4 & 67.5 & \second{94.8} & \second{94.4} & 79.8 & 76.4 & 86.7 & 83.1 & 65.0 \\

PPO~\citep{schulman2017proximal}
& RL
& 64.8 & 40.5 & 57.1 & 60.6 & 46.4 & 47.4 & 54.4 & 73.8 & 51.5 \\

RLOO~\citep{ahmadian2024back}
& RL
& 88.3 & 52.8 & 71.0 & 62.8 & 66.4 & 56.9 & 69.7 & 73.9 & 52.1 \\

Skill-GRPO$^\ast$~\citep{xia2026skillrl}
& RL
& 66.7 & 63.6 & \best{100.0} & \best{100.0} & 75.0 & \best{100.0} & 81.2 & \second{85.2} & 67.9 \\

OPD-\tch{Skill-1.5B}~\citep{ye2026policy}
& Distillation
& 85.7 & 46.2 & 73.1 & 55.6 & 76.7 & 53.3 & 71.1 & 84.7 & 69.1 \\

OPD-\tch{3B}~\citep{ye2026policy}
& Distillation
& 64.7 & 7.1 & 21.9 & 0.0 & 9.1 & 5.6 & 25.8 & 32.5 & 3.9\\

OPSD~\citep{zhao2026self}
& Distillation
& 26.3 & 16.7 & 9.1 & 6.7 & 9.1 & 5.3 & 14.1 & 22.3 & 10.2 \\

GRPO+OPD
& Hybrid
& \best{96.4} & \second{88.9} & 89.3 & 90.5 & 73.7 & \second{82.6} & \second{87.5} & 82.5 & \second{69.9} \\

GRPO+OPSD~\citep{lu2026self}
& Hybrid
& 87.1 & 66.7 & 85.7 & 58.3 & \best{85.0} & 46.2 & 72.7 & 79.9 & 66.8 \\

\textbf{\method{}}
& Hybrid
& \second{94.9} & \best{100.0} & 90.0 & \best{100.0} & \second{84.2} & 77.3 & \best{89.8} & \best{86.6} & \best{75.8} \\

\midrule
\rowcolor{gray!15}
\multicolumn{11}{l}{\itshape Qwen2.5-3B-Instruct} \\

Vanilla~\citep{qwen2.5}
& Prompting
& 44.4 & 11.1 & 6.2 & 15.4 & 28.6 & 12.5 & 21.9 & 6.7 & 0.8 \\

Skill-Prompt$^\ast$~\citep{xia2026skillrl}
& Prompting
& 51.7 & 66.7 & 48.4 & 0.0 & 4.3 & 10.0 & 28.9 & 0.2 & 0.8 \\

GRPO~\citep{shao2024deepseekmath}
& RL
& 91.2 & 62.5 & 96.2 & 61.9 & 65.0 & 47.4 & 75.0 & 79.8 & 63.3 \\

GiGPO~\citep{feng2026group}
& RL
& 95.8 & 81.8 & \second{96.4} & \best{100.0} & \second{88.9} & \best{86.4} & \second{92.2} & 83.1 & 69.5 \\

PPO~\citep{schulman2017proximal}
& RL
& 93.5 & 76.5 & 90.5 & 50.0 & 81.2 & 60.0 & 81.2 & 76.0 & 59.4 \\

RLOO~\citep{ahmadian2024back}
& RL
& 87.5 & 55.6 & 71.4 & 54.5 & 59.3 & 80.0 & 72.7 & 81.8 & 68.0 \\

Skill-GRPO$^\ast$~\citep{xia2026skillrl}
& RL
& 88.9 & 76.9 & \best{100.0} & 77.8 & 74.3 & 62.5 & 79.7 & 79.7 & 64.8 \\

OPD-\tch{Skill-3B}~\citep{ye2026policy}
& Distillation
& 74.3 & 80.0 & 76.9 & 69.2 & 81.8 & 58.8 & 74.2 & 79.1 & 66.4 \\

OPD-\tch{7B}~\citep{ye2026policy}
& Distillation
& 67.5 & 33.3 & 37.5 & 43.8 & 19.0 & 7.7 & 38.3 & 10.4 & 2.3 \\

OPSD~\citep{zhao2026self}
& Distillation
& 48.8 & 41.7 & 16.7 & 0.0 & 15.8 & 16.7 & 28.1 & 11.3 & 3.1 \\

GRPO+OPD
& Hybrid
& \second{96.6} & \best{90.0} & 83.3 & 87.5 & 66.7 & 78.9 & 82.8 & \second{84.5} & \second{74.2} \\

GRPO+OPSD~\citep{lu2026self}
& Hybrid
& \best{97.6} & 58.3 & 91.7 & \second{92.3} & 63.6 & 31.2 & 78.1 & 77.8 & 66.4 \\

\textbf{\method{}}
& Hybrid
& \best{97.6} & \second{83.3} & \best{100.0} & 85.7 & \best{95.5} & \second{83.3} & \best{93.8} & \best{85.8} & \best{77.3} \\

\midrule
\rowcolor{gray!15}
\multicolumn{11}{l}{\itshape Qwen2.5-7B-Instruct} \\

Vanilla~\citep{qwen2.5}
& Prompting
& 36.1 & 22.2 & 3.1 & 0.0 & 0.0 & 0.0 & 12.5 & 5.9 & 1.6 \\

Skill-Prompt$^\ast$~\citep{xia2026skillrl}
& Prompting
& 51.7 & 50.0 & 32.3 & 5.3 & 4.3 & 0.0 & 23.4 & 1.7 & 0.8 \\

GRPO~\citep{shao2024deepseekmath}
& RL
& 91.2 & \second{87.5} & 96.2 & 81.0 & 65.0 & 57.9 & 81.2 & 80.9 & 72.6 \\

GiGPO~\citep{feng2026group}
& RL
& \second{97.7} & 82.7 & \second{98.8} & 83.7 & \second{89.3} & 79.2 & 90.8 & 84.4 & 72.8 \\

PPO~\citep{schulman2017proximal}
& RL
& 92.3 & 64.0 & 92.5 & 89.5 & 80.3 & 68.8 & 80.4 & 81.4 & 68.7 \\

RLOO~\citep{ahmadian2024back}
& RL
& 87.6 & 78.2 & 87.3 & 81.3 & 71.9 & 48.9 & 75.5 & 80.3 & 65.7 \\

Skill-GRPO$^\ast$~\citep{xia2026skillrl}
& RL
& 97.1 & \best{100.0} & 92.3 & \best{100.0} & 85.0 & 72.7 & 90.6 & 88.7 & 78.9 \\

OPD-\tch{Skill-7B}~\citep{ye2026policy}
& Distillation
& \best{100.0} & 77.8 & 76.0 & \best{100.0} & 72.2 & \second{90.5} & 86.7 & 88.9 & 78.1 \\

OPD-\tch{14B}~\citep{ye2026policy}
& Distillation
& 84.2 & 40.0 & 77.8 & 42.1 & 66.7 & 47.4 & 64.8 & 28.8 & 4.7 \\

OPSD~\citep{zhao2026self}
& Distillation
& 50.0 & 60.0 & 22.7 & 21.4 & 17.6 & 9.5 & 32.8 & 4.5 & 2.3 \\

GRPO+OPD
& Hybrid
& \best{100.0} & 76.9 & 95.2 & \best{100.0} & 82.6 & 87.5 & \second{92.2} & \second{90.3} & \second{81.2} \\

GRPO+OPSD~\citep{lu2026self}
& Hybrid
& 94.3 & 58.8 & 81.0 & \second{94.4} & 88.2 & 50.0 & 79.7 & 86.8 & 76.5 \\

\textbf{\method{}}
& Hybrid
& \best{100.0} & \best{100.0} & \best{100.0} & 76.9 & \best{96.4} & \best{90.9} & \best{95.3} & \best{91.7} & \best{84.4} \\

\bottomrule
\end{tabular}%
}
\end{center}
\end{table*}

\subsection{Main Results}

\paragraph{Necessity of Skilled Teacher Training.} Explicitly training the teacher before distillation brings substantial gains over directly applying OPSD, highlighting the importance of effective skill utilization. As shown in Table~\ref{tab:main-results}, using a separately trained teacher for OPD instead of OPSD leads to a 14.8 points improvement on ALFWorld (87.5\% vs. 72.7\%) and a 3.1 points improvement on WebShop (69.9\% vs. 66.8\%) with Qwen2.5-1.5B-Instruct. These results demonstrate the necessity of training the teacher to effectively leverage its privileged context before using it for distillation.

\paragraph{Effectiveness beyond Pure RL and Distillation.} 
\method{} consistently improves over both pure reinforcement learning and distillation-based approaches across tasks and model scales. On ALFWorld, \method{} achieves a 93.8\% success rate with Qwen2.5-3B-Instruct, outperforming GRPO (75.0\%) and OPD (74.2\%). On WebShop, it reaches 75.8\% accuracy with Qwen2.5-1.5B-Instruct, compared with 56.8\% and 69.1\% for GRPO and OPD, respectively. Distillation-based methods remain constrained by fixed teacher supervision: OPD with a skilled teacher achieves strong performance but remains bounded by the fixed distillation process. These results show that \method{} effectively combines the complementary strengths of reinforcement learning and distillation while mitigating the limitations of fixed teacher supervision.

 \paragraph{Advantages over Naive Hybrid Approaches.}
Simply combining GRPO and OPD with a fixed objective is insufficient to capture their different roles throughout training. \method{} improves over the GRPO+OPD baseline by 11.0 points on ALFWorld and 3.1 points on WebShop with the 3B model. This highlights the importance of adaptive teacher retirement, which dynamically adjusts the reliance on teacher supervision as the student improves.

\subsection{Training Dynamics}
\label{sec:dynamics}
\begin{figure}[t]

\begin{center}
\includegraphics[
    width=\linewidth,
    trim={0 0 0 0},
    clip
]{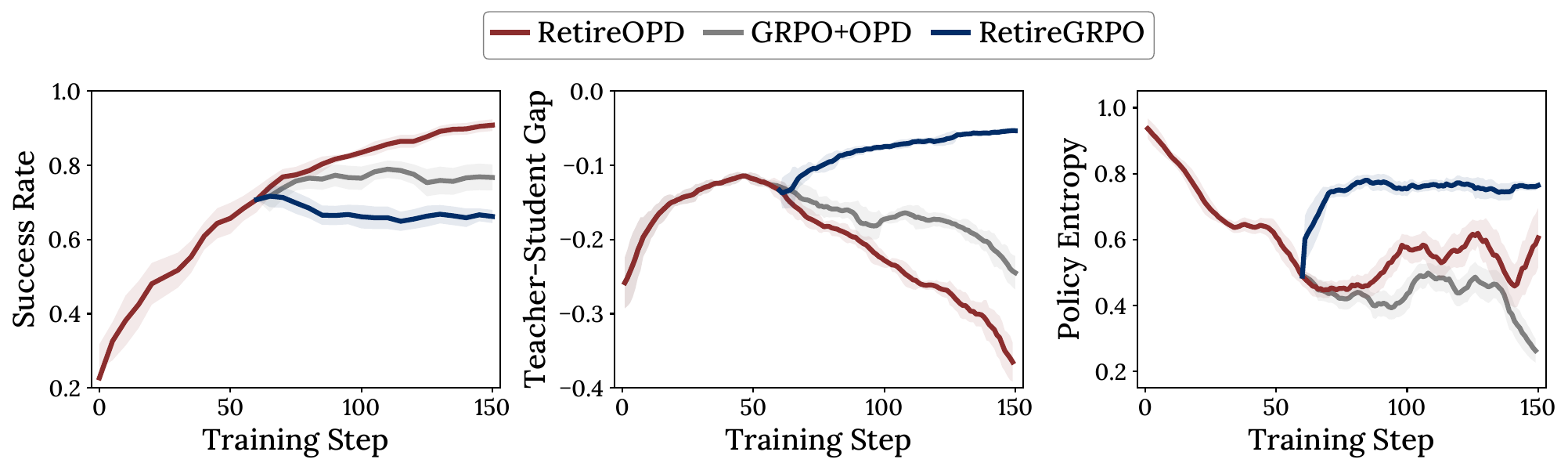}
\end{center}

\caption{\textbf{Left:} Success rate of student model on ALFWorld. \textbf{Middle:} Teacher-student gap on ALFWorld. \textbf{Right:} Entropy of student model on ALFWorld.}
\label{fig:exit_comparison}

\end{figure}

\paragraph{Performance Trends.} The left panel of Figure~\ref{fig:exit_comparison} illustrates the training dynamics of GRPO+OPD, RetireGRPO, and \method{} on Qwen2.5-3B-Instruct. RetireGRPO denotes the variant that adaptively removes GRPO while retaining OPD. GRPO+OPD converges rapidly after around 60 steps, while RetireGRPO suffers a performance drop after removing GRPO, as OPD drives the student toward the teacher and limits further improvement. In contrast, \method{} continues to improve after adaptively removing OPD, indicating that timely retirement can release the student from the teacher's capability ceiling and enable further RL-driven improvement.

\paragraph{Teacher–Student Alignment and Exploration.} We further analyze the teacher-student gap and entropy (Figure~\ref{fig:exit_comparison} middle and right). GRPO+OPD continuously pulls the student toward the teacher, while RetireGRPO allows the student to move even closer to the teacher. In contrast, our method avoids excessive teacher fitting, preserving the student's capacity for improvement. The entropy trends further support this observation: removing GRPO leads to increased entropy and less stable training, whereas GRPO+OPD drives entropy downward, indicating reduced exploration. Our method maintains entropy at a moderate level, balancing training stability and exploration.

\subsection{Ablation Studies}
\label{sec:ablation}

\begin{figure*}[h]
  \centering

  \begin{minipage}[t]{0.49\textwidth}
    \centering
    \includegraphics[
      width=\linewidth,
      trim={0 0 388.8bp 0},
      clip
    ]{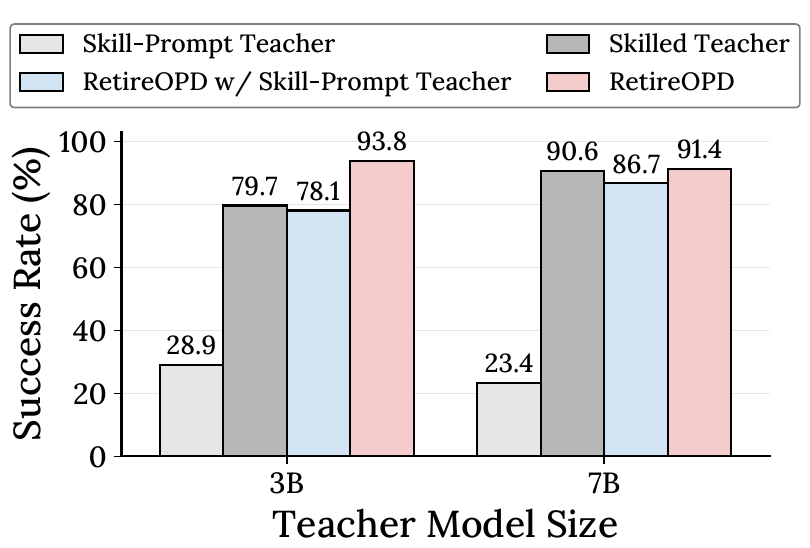}
    \captionof{figure}{
      Ablation study of teacher construction on ALFWorld.
    }
    \label{fig:teacher_ablation}
  \end{minipage}
  \hfill
  \begin{minipage}[t]{0.49\textwidth}
    \centering
    \includegraphics[
      width=\linewidth,
      trim={388.8bp 0 0 0},
      clip
    ]{teacher_construction_two_models_with_atod_side_by_side.pdf}
    \captionof{figure}{
      Adaptive Retirement vs. Linear Annealing on ALFWorld.
    }
    \label{fig:atod_comparison}
  \end{minipage}

\end{figure*}

\paragraph{Teacher Construction.} \label{ablation:teacher_construction}
Figure~\ref{fig:teacher_ablation} compares teachers with and without skill training, together with the corresponding Qwen2.5-3B-Instruct students trained with our method. Dedicated training substantially improves the teacher, with consistent gains across both model sizes. More importantly, the improved teacher provides more effective supervision for the student. For the 3B teacher, the student reaches 93.8\% when trained with the dedicated skill-teacher, compared with 78.1\% when using the untrained teacher, while the 7B teacher yields a similar improvement from 86.7\% to 91.4\%. These results demonstrate that dedicated skill training not only enhances the teacher's ability to exploit privileged context, but also enables more effective supervision for the student.

\paragraph{Retirement Criteria.} Table~\ref{tab:exit_ablation} presents an ablation of the retirement mechanism on Qwen2.5-3B-Instruct on ALFWorld. Removing either Alignment Stagnation or Relative Competence reduces the success rate from 93.8\% to 89.0\%, showing that both components contribute to effective retirement. Removing retirement entirely further reduces performance to 82.8\%, while removing OPD from the beginning and using GRPO alone achieves only 75.0\%. These results show that both early-stage distillation and later-stage independent RL optimization are important, and that \method{} benefits from adaptively determining when to retire OPD.

% =========================
% Module ablations
% =========================
\begin{table}[t]
\centering

% ==================== Left Table ====================
\begin{minipage}[t]{0.48\linewidth}
\centering
\caption{Retirement mechanism ablation.}
\label{tab:exit_ablation}

\small
\renewcommand{\arraystretch}{1.05}
\setlength{\tabcolsep}{4pt}

\begin{tabularx}{0.92\linewidth}
{@{}>{\raggedright\arraybackslash}Xcc@{}}
\toprule
\textbf{Setting} & \textbf{Retire @} & \textbf{SR} \\
\midrule

\rowcolor{gray!15}
\textbf{\method{} (Ours)} & \textbf{60} & \textbf{93.8} \\

\quad w/o Rebound Indicator     & 50  & 89.0 \\
\quad w/o Relative Competence      & 100 & 89.0 \\
\quad w/o Retirement            & --  & 82.8 \\
\quad w/o OPD (GRPO-only) & 0   & 75.0 \\

\bottomrule
\end{tabularx}
\end{minipage}
\hfill
% ==================== Right Table ====================
\begin{minipage}[t]{0.48\linewidth}
\centering
% \caption{Parameter ablation.}
\caption{Retirement threshold ablation.}
\label{tab:parameter_sensitivity}

\small
\renewcommand{\arraystretch}{1.05}
\setlength{\tabcolsep}{4pt}

% \begin{tabularx}{0.9\linewidth}
% {@{}>{\raggedright\arraybackslash}Xcc@{}}
% \toprule
% \textbf{Setting} & \textbf{Retire @} & \textbf{SR} \\
% \midrule

% \rowcolor{gray!15}
% $\gamma=0.9,\ \delta=0$ \textbf{(Ours)}
% & \textbf{60} & \textbf{93.8} \\

% $\gamma=0.8,\ \delta=0$
% & 55 & 90.6 \\

% $\gamma=1.0,\ \delta=0$
% & 95 & 91.4 \\

% $\gamma=0.9,\ \delta=0.03$
% & 65 & 89.1 \\

% $\gamma=0.9,\ \delta=0.05$
% & 95 & 91.4 \\

% \bottomrule
% \end{tabularx}

\begin{tabularx}{0.9\linewidth}
{@{}>{\raggedright\arraybackslash}Xcc@{}}
\toprule
\textbf{Setting} & \textbf{Retire @} & \textbf{SR} \\
\midrule

\rowcolor{gray!15}
$\delta=0,\ \gamma=0.9$ \textbf{(Ours)}
& \textbf{60} & \textbf{93.8} \\

$\delta=0.03,\ \gamma=0.9$
& 65 & 89.1 \\

$\delta=0.05,\ \gamma=0.9$
& 75 & 91.4 \\

$\delta=0,\ \gamma=0.8$
& 55 & 90.6 \\

$\delta=0,\ \gamma=1.0$
& 95 & 91.4 \\

\bottomrule
\end{tabularx}

\end{minipage}

\end{table}

\paragraph{Sensitivity to Retirement Thresholds.} We analyze how the thresholds $\gamma$ and $\delta$ affect the retirement timing. As shown in Table~\ref{tab:parameter_sensitivity}, increasing either threshold generally leads to a later retirement point. A larger $\delta$ requires a more pronounced stagnation or rebound of the alignment gap, while a larger $\gamma$ requires a larger relative competence ratio before retirement. Despite these changes in retirement timing, the final performance remains relatively stable across different threshold values, indicating that \method{} is not highly sensitive to the choice of thresholds.

\paragraph{Adaptive Retirement vs. Linear Annealing.} We further compare our method with a Linear Annealing~\citep{tan2026atod} strategy, which replaces adaptive retirement with a predefined linear schedule (Figure~\ref{fig:atod_comparison}). \method{} exhibits a more stable training trajectory, with performance continuing to improve throughout training, whereas Linear Annealing shows pronounced fluctuations. It also achieves a higher final success rate, demonstrating that adaptively determining when to retire OPD is more effective than relying on a fixed linear schedule.

\begin{figure}[h]

\begin{center}
\includegraphics[width=1\linewidth]{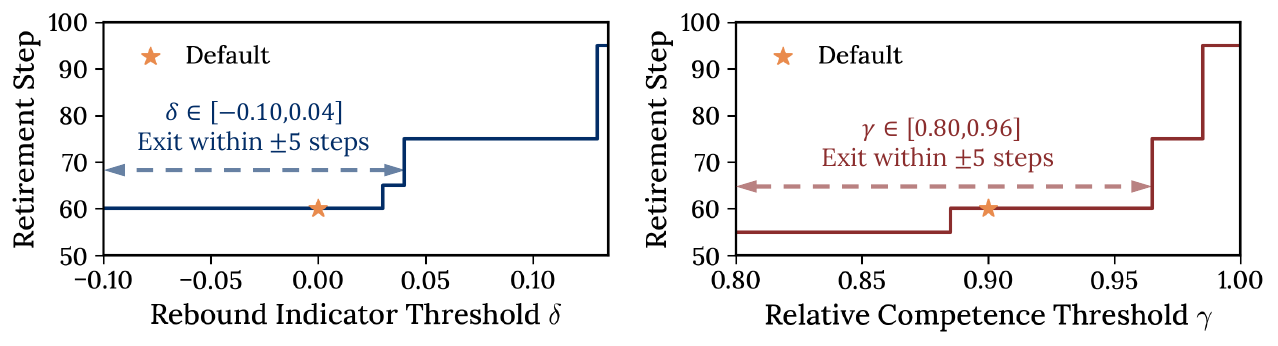}
\end{center}

\caption{Robustness of the retirement mechanism to threshold variations. \textbf{Left:} Sensitivity to the rebound indicator threshold $\delta$. \textbf{Right:} Sensitivity to the relative competence threshold $\gamma$.}
\label{fig:retire_timing}

\end{figure}

\subsection{Robustness to Retirement Thresholds} Figure~\ref{fig:retire_timing} examines the robustness of the retirement point to variations in $\delta$ and $\gamma$. Across a relatively wide range of values, the retirement point remains stable: varying $\delta$ from -0.1 to 0.04 results in a similarly small variation of around 5 steps, while varying $\gamma$ from 0.8 to 0.96 changes the retirement step by at most $\pm5$ steps. This stability indicates that \method{} is robust to the choice of retirement thresholds, with the adaptive mechanism consistently identifying a similar transition point despite substantial changes in the threshold values.

\section{Conclusion}
% We proposed \method{}, which internalizes privileged skills into LLM agents without requiring such information at inference time. It first optimizes a skill-conditioned teacher with environment rewards, then trains a skill-free student jointly with GRPO and on-policy distillation, and adaptively retires teacher supervision once behavioral transfer stagnates and the student reaches sufficient task competence. Across three model scales on ALFWorld and WebShop, \method{} consistently improves performance and outperforms pure reinforcement learning, continuous joint optimization of GRPO and OPD and the corresponding skill-conditioned teachers.

We propose \method{} based on two observations: a skill-conditioned teacher without dedicated training may fail to exploit privileged skills, while continued on-policy distillation can constrain the student's performance by the teacher's capability ceiling. \method{} first trains a dedicated skill-conditioned teacher, then uses it to train a skill-free student, with the adaptive retirement mechanism that removes teacher supervision during training, freeing the student from the teacher's capability constraints. Experimental results show that our method consistently outperforms pure reinforcement learning, distillation-based, and hybrid approaches, validating its ability to effectively transfer privileged skills while avoiding the limitations imposed by continued teacher supervision.

\bibliography{ref}
\bibliographystyle{iclr2027_conference}

\clearpage
\appendix
\section{Appendix}
\subsection{Theoretical Analysis}

\label{sec:theory}
\paragraph{Setup.}
We analyze GRPO-OPD interaction through a local teacher-regularized view.
Let
\begin{gather*}
J(\theta)
=
\mathbb{E}_{x,\tau\sim\pi_\theta}[R(\tau)],\\
D(\theta)
=
\mathbb{E}_{x,y\sim\pi_\theta}
\left[
\frac{1}{|y|}
\sum_{t=1}^{|y|}
D_{\mathrm{KL}}
\bigl(
\pi_\theta(\cdot\mid x,y_{<t})
\Vert
\pi_T(\cdot\mid x,c^+,y_{<t})
\bigr)
\right],\\
F_\lambda(\theta)
=
J(\theta)-\lambda D(\theta),
\qquad
g=\nabla_\theta J,
\qquad
h=\nabla_\theta D.
\end{gather*}
Ignoring finite-sample noise and assuming inactive GRPO clipping locally,
the expected update is
\begin{equation*}
\theta^+
=
\theta+\eta(g-\lambda h),
\end{equation*}
where $g$ is the reward-driven direction and $-h$ reduces the
teacher--student discrepancy.

\paragraph{Interaction between GRPO and OPD.}
A first-order expansion gives
\begin{align*}
J(\theta^+)-J(\theta)
&=
\left\langle
\nabla_\theta J(\theta),
\theta^+-\theta
\right\rangle
+O(\eta^2)\\
&=
\eta\left\langle
g,\,g-\lambda h
\right\rangle
+O(\eta^2)\\
&=
\eta
\left(
\|g\|^2-\lambda\langle g,h\rangle
\right)
+O(\eta^2),\\[1mm]
\langle g,h\rangle<0
&\Rightarrow
\text{OPD accelerates reward optimization},\\
\langle g,h\rangle>0
&\Rightarrow
\text{OPD conflicts with reward optimization}.
\end{align*}
Moreover, if
$\lambda\langle g,h\rangle>\|g\|^2$,
the joint update decreases the expected return to first order.

\paragraph{Discrepancy as a Conflict Signal.}
Under joint optimization,
\begin{align*}
\frac{\mathrm d}{\mathrm dt}D(\theta)
&=
\left\langle
\nabla_\theta D(\theta),
\frac{\mathrm d\theta}{\mathrm dt}
\right\rangle \\
&=
\left\langle
h,\, g-\lambda h
\right\rangle \\
&=
\langle g,h\rangle-\lambda\|h\|^2.
\end{align*}
Thus, a non-decreasing discrepancy indicates local conflict between
the GRPO direction $g$ and the OPD direction $-h$. If $h\approx0$,
OPD already provides little additional first-order guidance. Discrepancy
stagnation therefore signals either emerging conflict or a diminishing
OPD signal.

\paragraph{Benefit of Teacher Retirement.}
Suppose GRPO and OPD are locally conflicting, i.e.,
$\langle g,h\rangle>0$. Consider
\begin{align*}
\theta_{\mathrm{joint}}^+
&=
\theta+\eta(g-\lambda h),\\
\theta_{\mathrm{retirement}}^+
&=
\theta+\eta g.
\end{align*}
Their expected returns satisfy
\begin{align*}
J(\theta_{\mathrm{retirement}}^+)
-
J(\theta_{\mathrm{joint}}^+)
&=
\Bigl[
J(\theta)+\eta\|g\|^2
\Bigr] \\
&\quad -
\Bigl[
J(\theta)
+\eta\bigl(\|g\|^2-\lambda\langle g,h\rangle\bigr)
\Bigr]
+O(\eta^2)\\
&=
\eta\lambda\langle g,h\rangle
+O(\eta^2).
\end{align*}
Since $\langle g,h\rangle>0$, teacher retirement yields a larger
first-order reward improvement than continued joint optimization for
sufficiently small $\eta$.

\paragraph{Implication for Adaptive Retirement.}
The analysis shows that OPD accelerates learning while teacher guidance aligns with reward optimization, whereas stagnation of the teacher-student discrepancy indicates diminishing OPD benefit or emerging conflict with GRPO. We therefore use stagnation of $\bar K_m$ as the retiring signal and additionally require sufficient student competence to avoid premature retirement caused by early training noise.
\clearpage
\subsection{Algorithm}
\label{app:algorithm}

The complete training procedure of our method, including privileged
teacher construction, joint GRPO-OPD training, and adaptive teacher
retirement, is summarized in Algorithm~\ref{alg:\method{}}.

\begin{algorithm}[H]
\caption{\textsc{\method{}} Training Pipeline}
\label{alg:\method{}}
\begin{algorithmic}[1]

\Require Base policy $\pi_0$;
training set $\mathcal{D}=\{(x,c^+)\}$;
validation set $\mathcal{D}_{\mathrm{val}}$;
environment reward $R$;
group size $G$;
OPD weight $\lambda$;
monitoring window $W$;
exit thresholds $\gamma,\delta$

% =========================================================
% Phase 1
% =========================================================
\Statex
\Statex \(\triangleright\) \textbf{Phase 1: Privileged Teacher Construction}

\State Initialize teacher policy $\pi_\phi\gets\pi_0$

\For{each teacher training iteration}

    \State Sample $(x,c^+)\sim\mathcal{D}$

    \State
    $\{y^{(i)}\}_{i=1}^{G}
    \sim\pi_\phi(\cdot\mid x,c^+);
    \quad
    R^{(i)}\gets R(x,y^{(i)})$

    \State
    $\hat A^{(i)}
    \gets
    \dfrac{R^{(i)}-\mu_G}{\sigma_G},
    \qquad
    r^{(i)}_{t}
    \gets
    \dfrac{
    \pi_\phi(y^{(i)}_t\mid x,c^+,y^{(i)}_{<t})
    }{
    \pi_{\phi_{\mathrm{old}}}(y^{(i)}_t\mid x,c^+,y^{(i)}_{<t})
    }$

    \State
    Update $\phi$ by minimizing
    $\mathcal{L}_{\mathrm{GRPO}}
    (\phi;\{\hat A^{(i)},r^{(i)}_{t}\})$

\EndFor

\State
$\pi_T\gets\operatorname{StopGrad}(\pi_\phi);
\qquad
SR_T\gets
\operatorname{Success}
(\pi_T,\mathcal{D}_{\mathrm{val}};c^+)$

% =========================================================
% Phase 2
% =========================================================
\Statex
\Statex \(\triangleright\) \textbf{Phase 2: Joint GRPO-OPD with Adaptive Retirement}

\State
$\pi_\theta\gets\pi_0;
\quad
\textsc{Retired}\gets\textbf{false};
\quad
m\gets0$

\For{each student training iteration $n$}

    \State Sample $(x,c^+)\sim\mathcal{D}$

    \State
    $\{y^{(i)}\}_{i=1}^{G}
    \sim\pi_\theta(\cdot\mid x);
    \quad
    R^{(i)}\gets R(x,y^{(i)})$
    \quad\textcolor{algorithmblue}{\(\triangleright\) \textbf{Skill-free on-policy rollout}}

    \State
    $\hat A^{(i)}
    \gets
    \dfrac{R^{(i)}-\mu_G}{\sigma_G},
    \qquad
    r^{(i)}_{t}
    \gets
    \dfrac{
    \pi_\theta(y_{t}^{(i)}\mid x,y_{<t}^{(i)})
    }{
    \pi_{\theta_{\mathrm{old}}}(y_{t}^{(i)}\mid x,y_{<t}^{(i)})
    }$
    \quad\textcolor{algorithmblue}{\(\triangleright\) \textbf{GRPO objective}}

    \If{\textbf{not} $\textsc{Retired}$}

        \State
        $\Delta_{t}^{(i)}
        \gets
        \log\pi_T(y_{t}^{(i)}\mid x,c^+,y_{<t}^{(i)})
        -
        \log\pi_\theta(y_{t}^{(i)}\mid x,y_{<t}^{(i)})$

        \State
        $K_n
        \gets
        \dfrac{1}{G}
        \sum_{i=1}^{G}
        \dfrac{1}{|y^{(i)}|}
        \sum_{t=1}^{|y^{(i)}|}
        \Delta_{t}^{(i)};
        \qquad
        \mathcal{L}_{\mathrm{OPD}}\gets -K_n$
        \quad\textcolor{algorithmblue}{\(\triangleright\) \textbf{On-policy distillation}}

        \State
        Update $\theta$ by minimizing
        $\mathcal{L}_{\mathrm{GRPO}}
        +
        \lambda\mathcal{L}_{\mathrm{OPD}}$

    \Else

        \State
        Update $\theta$ by minimizing
        $\mathcal{L}_{\mathrm{GRPO}}$

    \EndIf

    % =====================================================
    % Adaptive Exit
    % =====================================================

    \If{$n\bmod W=0$ \textbf{ and not} $\textsc{Retired}$}
        \State
        $m\gets m+1;
        \qquad
        \bar K_m
        \gets
        \dfrac{1}{W}
        \sum_{n=mW}^{mW+W-1}K_n$

        \State
        $SR_m
        \gets
        \operatorname{Success}
        (\pi_\theta,\mathcal{D}_{\mathrm{val}})$

        \If{$m\ge2$}

            \State
            $\rho_m^{(K)}
            \gets
            \dfrac{\bar K_m-\bar K_{m-1}}{\bar K_m};
            \qquad
            \eta_m
            \gets
            \dfrac{SR_m+SR_{m-1}}{2SR_T}$

            \If{
            $\rho_m^{(K)}\ge\delta$
            \textbf{ and }
            $\eta_m\ge\gamma$
            }
            \quad\textcolor{algorithmblue}{\(\triangleright\) \textbf{Adaptive teacher retirement}}
                \State
                $\textsc{Retired}\gets\textbf{true}$

            \EndIf

        \EndIf

    \EndIf

\EndFor

\State
\textbf{\Return} skill-free student policy $\pi_\theta$

\end{algorithmic}
\end{algorithm}

\subsection{Hyperparameter Settings}
\label{app:detail}

We implement all methods under a unified training setup to ensure a 
consistent comparison. Table~\ref{tab:hyperparameters} summarizes the 
optimization, data, and method-specific hyperparameters used in our 
experiments. Methods with identical configurations are grouped into the 
same column for compact presentation. $G$ denotes the group size, $\lambda$ is the distillation loss coefficient,
and $\alpha_{\mathrm{KL}}$ is the KL-divergence penalty coefficient.
OP(S)D collectively denotes OPD and OPSD, while GRPO+OP(S)D denotes
GRPO+OPD and GRPO+OPSD. GRPO, GiGPO and RLOO are grouped because they
use the same hyperparameter configuration. 

\begin{table*}[h]
\centering
\caption{Hyperparameter settings.
A dash indicates that the
corresponding hyperparameter is not applicable.}
\label{tab:hyperparameters}
\setlength{\tabcolsep}{4pt}
\resizebox{\linewidth}{!}{%
\begin{tabular}{lccccc}
\toprule
& \multicolumn{5}{c}{\textbf{Method}} \\
\cmidrule(lr){2-6}
\textbf{Hyperparameter}
& \textbf{\method{}}
& \textbf{GRPO / GiGPO / RLOO}
& \textbf{PPO}
& \textbf{OP(S)D}
& \textbf{GRPO+OP(S)D} \\
\midrule

Learning rate
& $1{\times}10^{-6}$
& $1{\times}10^{-6}$
& $1{\times}10^{-6}$
& $1{\times}10^{-6}$
& $1{\times}10^{-6}$ \\

Critic learning rate
& -- & -- & $1{\times}10^{-5}$ & -- & -- \\

$G$
& 8 & 8 & -- & 8 & 8 \\

$\lambda$
& 0.01 & -- & -- & 0.01 & 0.01 \\

$\alpha_{\mathrm{KL}}$
& 0.01 & 0.01 & 0.01 & 0.01 & 0.01 \\

Train batch size
& 16 & 16 & 16 & 16 & 16 \\

Validation data size
& \multicolumn{5}{c}{128 (ALFWorld) / 256 (WebShop)} \\

Competence threshold
& 0.9 & -- & -- & -- & -- \\

Window size
& 5 & -- & -- & -- & -- \\

Training steps
& 150 & 150 & 150 & 150 & 150 \\

\bottomrule
\end{tabular}%
}
\end{table*}

% \paragraph{ATOD Linear Annealing}
% \label{app:annealing}

% Except for the loss-weight schedule, all experimental settings of ATOD are identical to those of our main method. The training objective is defined as
% \begin{equation*}
%     \mathcal{L}_{\mathrm{ATOD}}(s)
%     =
%     \lambda_{\mathrm{RL}}(s)\mathcal{L}_{\mathrm{GRPO}}
%     +
%     \lambda_{\mathrm{KL}}(s)\mathcal{L}_{\mathrm{OPD}},
% \end{equation*}
% where $s$ denotes the current training step. We define the annealing progress as
% \begin{equation*}
%     p(s)=\min\left(\frac{s}{80},1\right).
% \end{equation*}
% The two loss weights are scheduled as
% \begin{equation*}
%     \lambda_{\mathrm{KL}}(s)=1-0.9p(s),
%     \qquad
%     \lambda_{\mathrm{RL}}(s)=0.1+p(s).
% \end{equation*}

% At the beginning of training, the GRPO and OPD weights are $0.1$ and $1.0$, respectively, such that optimization is primarily guided by the teacher distillation signal. As training proceeds, the GRPO weight gradually increases while the OPD weight decreases. From step $80$ onward, the two weights remain fixed at $1.1$ and $0.1$, respectively. This schedule smoothly shifts the optimization from OPD-dominated learning to reward-driven GRPO while retaining a small amount of distillation regularization.
\subsection{Analysis of Teacher Restoration after Retirement}

\begin{wrapfigure}{r}{0.45\textwidth}
    \vspace{-10pt}
    \centering
    \includegraphics[width=\linewidth]
    {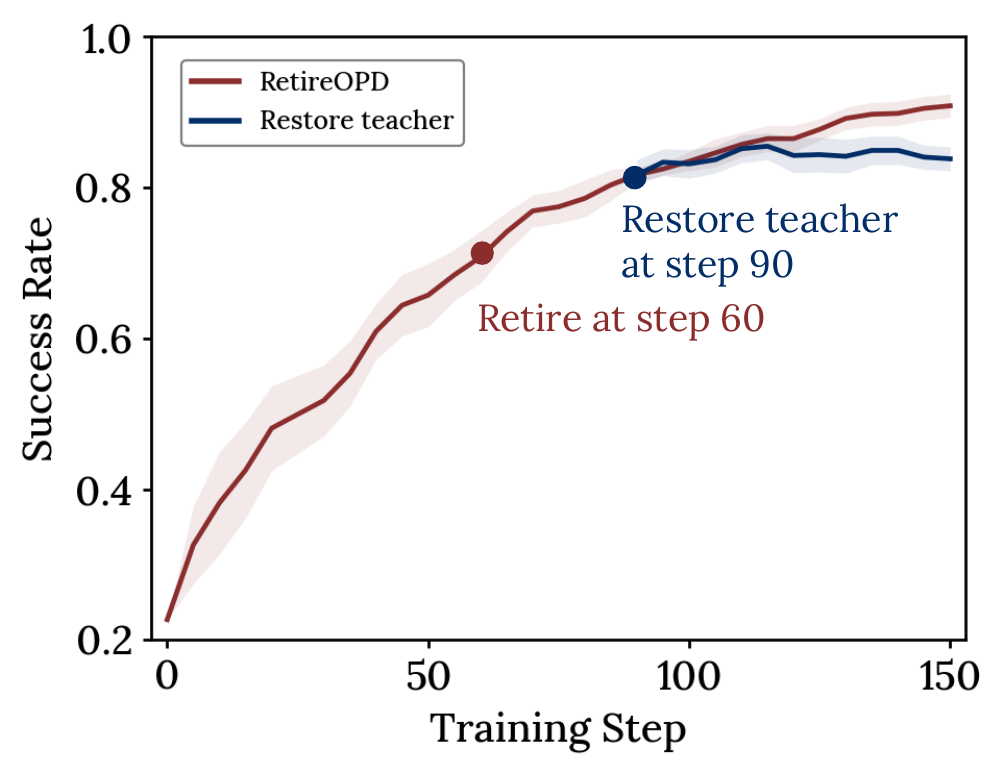}
    \caption{Teacher restoration after retirement.}
    \label{fig:teacher_restore}
    \vspace{-10pt}
\end{wrapfigure}

\paragraph{Does Teacher Supervision Become Useful Again?}
We further examine whether the need for teacher supervision may re-emerge after retirement. 
After the retirement criterion is triggered at step 60, we continue reward-only training and restore teacher supervision at step 90. 
As shown in Figure~\ref{fig:teacher_restore}, restoring the teacher does not yield additional improvement, whereas the model that keeps the teacher retired continues to improve and eventually achieves a higher success rate. 
This observation suggests that, in this setting, the retirement criterion captures a relatively persistent change in the utility of teacher supervision rather than a short-lived fluctuation. 
Within the remaining training horizon, reintroducing the same teacher does not recover the benefit observed during the earlier stage of training.

\subsection{Details of Privileged Teacher Construction} \label{app:teacher_construction}

We provide additional details of the privileged teacher construction used throughout our experiments. 

\paragraph{Teacher Initialization and Skill Conditioning.} For each model scale, the teacher is initialized from the same instruction-tuned backbone as the student. The key difference is that the teacher additionally receives task-relevant privileged skill context $c^{+}$ throughout interaction. Specifically, at each decision step, the teacher conditions its action on the task input, interaction history and $c^{+}$, whereas the student never has access to such privileged information. The same skill format is used throughout teacher training and evaluation. 

\paragraph{Teacher Optimization.} We train the teacher using GRPO with privileged skills provided as part of its context, corresponding to the Skill-GRPO setting described in Section~\ref{sec:teacher_construction}. Apart from the additional skill context, teacher training follows the same environment interaction and reward definition used for student optimization. For each sampled group of trajectories, environment rewards are used to compute group-relative advantages and update the skill-conditioned policy. The teacher uses the same hyperparameter configuration as GRPO reported in Table~\ref{tab:hyperparameters}, including the group size, learning rate and generation settings. During evaluation, the teacher is also provided with the privileged skill context $c^{+}$, allowing us to directly assess its ability to leverage privileged information for task solving. We report this evaluation setting as Skill-GRPO$^\ast$ in Table~\ref{tab:main-results}.

\paragraph{Checkpoint Selection and Downstream Usage.} We periodically evaluate the teacher on the validation set and select the checkpoint according to the corresponding task metric. The selected teacher is then frozen and used throughout student training to provide token-level supervision. Its validation performance is also recorded as the teacher reference performance used by the adaptive retirement criterion. No teacher parameters are updated during student optimization. 

The effectiveness of explicitly optimizing the skill-conditioned teacher, compared with directly prompting the base model with the same privileged skills, is examined separately in the ablation study in Section~\ref{ablation:teacher_construction}.
\subsection{Ablation on OPD Weight}

\begin{wrapfigure}{r}{0.45\textwidth}
    \vspace{-10pt}
    \centering
    \includegraphics[width=\linewidth]
    {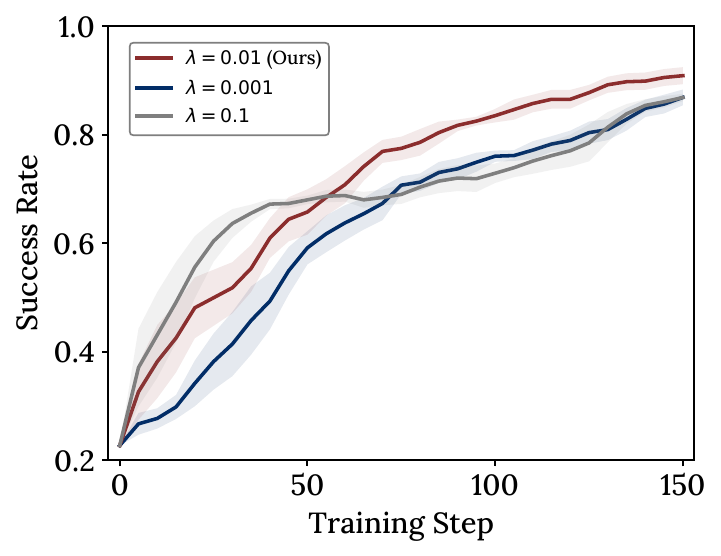}
    \caption{\textbf{Sensitivity to the distillation weight $\lambda$.}
    Success rate during training under different fixed distillation
    weights. Shaded regions indicate variability across runs.}
    \label{fig:lambda_sensitivity}
    \vspace{-10pt}
\end{wrapfigure}

% \paragraph{Sensitivity to Distillation Weight.}
Figure~\ref{fig:lambda_sensitivity} studies the effect of the
distillation weight $\lambda$. A relatively large weight
($\lambda=0.1$) leads to the fastest improvement during early
training, showing that stronger teacher supervision can effectively
facilitate initial policy learning. However, its improvement slows
down in later stages and results in a lower final success rate.
In contrast, a small weight ($\lambda=0.001$) provides weaker
guidance and therefore learns substantially more slowly at the
beginning, although it gradually catches up as training proceeds.
The default setting $\lambda=0.01$ achieves a better balance between
these two regimes, maintaining effective early-stage guidance while
allowing continued reward-driven improvement, and consequently
achieves the best final performance. These dynamics further
illustrate that the preferred strength of teacher supervision
changes over training, motivating an adaptive mechanism for
removing teacher guidance once it becomes less beneficial.

\subsection{Case Study}

To provide a more fine-grained view of how teacher supervision affects policy learning, we examine token-level teacher scores on representative ALFWorld trajectories. We focus on two complementary cases: how privileged skills improve supervision during early exploration, and how the same teacher supervision can become restrictive after the student has acquired sufficient task competence.

\paragraph{Skill-informed supervision during early exploration.}
Figure~\ref{fig:early-skilled-vs-opsd-clean} compares our skilled teacher with the OPSD teacher on the same student trajectory. While both assign similar scores to routine actions, clear differences emerge at decision-critical states requiring task-specific knowledge. At Step~8, the agent reaches the sink holding the target lettuce, and the retrieved skill specifies cleaning before placement. Our teacher assigns much higher likelihood to \texttt{clean} ($-0.826$) than the OPSD teacher ($-16.625$), showing that explicit teacher optimization better translates privileged skills into effective token-level supervision.

\begin{figure*}[h]
\centering
\small
\setlength{\tabcolsep}{3pt}
\renewcommand{\arraystretch}{1.08}

\begin{tabular}{r p{0.18\textwidth} p{0.18\textwidth} p{0.25\textwidth} p{0.25\textwidth}}
\toprule
Step & State summary & Retrieved skill & Skilled teacher score & OPSD teacher score \\
\midrule

0 &
initial search &
Clean before placing &
\tokcell{go}{0.000}
\tokcell{to}{0.000}
\tokcell{drawer}{0.000}
\tokcell{1}{0.000}
&
\tokcell{go}{-0.001}
\tokcell{to}{0.000}
\tokcell{drawer}{0.000}
\tokcell{1}{0.000}
\\

1 &
drawer search &
Search likely containers &
\tokcell{open}{0.000}
\tokcell{drawer}{0.000}
\tokcell{1}{0.000}
&
\tokcell{open}{0.000}
\tokcell{drawer}{0.000}
\tokcell{1}{0.000}
\\

2 &
cabinet search after empty drawer &
Search likely containers &
\tokcell{open}{-4.019}
\tokcell{cabinet}{-1.313}
\tokcell{10}{-0.017}
&
\tokcell{open}{-6.752}
\tokcell{cabinet}{-0.079}
\tokcell{10}{-0.033}
\\

3 &
moving to likely food surface &
Search likely food surfaces &
\tokcell{go}{-0.003}
\tokcell{to}{0.000}
\tokcell{diningtable}{-0.014}
\tokcell{1}{0.000}
&
\tokcell{go}{-0.001}
\tokcell{to}{0.000}
\tokcell{diningtable}{-0.011}
\tokcell{1}{0.000}
\\

4 &
lettuce found on dining table &
Pick target object &
\tokcell{take}{-0.003}
\tokcell{lettuce}{0.000}
\tokcell{2}{0.000}
\tokcell{from}{0.000}
\tokcell{diningtable}{0.000}
\tokcell{1}{0.000}
&
\tokcell{take}{-0.007}
\tokcell{lettuce}{0.000}
\tokcell{2}{0.000}
\tokcell{from}{0.000}
\tokcell{diningtable}{0.000}
\tokcell{1}{0.000}
\\

5 &
checking object state &
Verify object state &
\tokcell{examine}{-0.001}
\tokcell{lettuce}{0.000}
\tokcell{2}{0.000}
&
\tokcell{examine}{-0.003}
\tokcell{lettuce}{0.000}
\tokcell{2}{0.000}
\\

6 &
lettuce held; cleaning required &
Use sink for cleaning &
\tokcell{go}{-0.002}
\tokcell{to}{0.000}
\tokcell{sinkbasin}{0.000}
\tokcell{1}{0.000}
&
\tokcell{go}{-0.003}
\tokcell{to}{0.000}
\tokcell{sinkbasin}{0.000}
\tokcell{1}{0.000}
\\

7 &
before cleaning &
Verify object state &
\tokcell{examine}{-0.049}
\tokcell{lettuce}{0.000}
\tokcell{2}{0.000}
&
\tokcell{examine}{-0.049}
\tokcell{lettuce}{0.000}
\tokcell{2}{0.000}
\\

8 &
at sink; cleaning should happen &
Use sink for cleaning
Clean before placing &
\tokcell{clean}{-0.826}
\tokcell{lettuce}{0.000}
\tokcell{2}{0.000}
\tokcell{with}{0.000}
\tokcell{sinkbasin}{0.000}
\tokcell{1}{0.000}
&
\tokcell{clean}{-16.625}
\tokcell{lettuce}{0.000}
\tokcell{2}{0.000}
\tokcell{with}{0.000}
\tokcell{sinkbasin}{0.000}
\tokcell{1}{0.000}
\\

9 &
after cleaning; place target object &
Place after cleaning &
\tokcell{move}{-0.015}
\tokcell{lettuce}{0.000}
\tokcell{2}{0.000}
\tokcell{to}{0.000}
\tokcell{countertop}{-0.013}
\tokcell{1}{0.000}
&
\tokcell{move}{-0.013}
\tokcell{lettuce}{0.000}
\tokcell{2}{0.000}
\tokcell{to}{0.000}
\tokcell{countertop}{-0.576}
\tokcell{1}{0.000}
\\

\bottomrule
\end{tabular}

\vspace{2pt}
\caption{
\textbf{Early-stage teacher comparison on ALFWorld.} Both teachers score the same student-generated action tokens. The retrieved skill provides privileged guidance available only to the skilled teacher. At the critical cleaning step, the skilled teacher assigns substantially higher likelihood to the key token \texttt{clean}.}
\label{fig:early-skilled-vs-opsd-clean}
\end{figure*}

\paragraph{Teacher conflict after competence improves.} Figure~\ref{fig:teacher-token-conflict} shows a successful trajectory in which the student takes a task-effective action that is strongly disfavored by the teacher. In particular, the transition toward the microwave receives a low teacher score despite leading to successful task completion. This illustrates that, once the student becomes sufficiently competent, continued teacher matching can conflict with reward-driven optimization, motivating adaptive teacher exit.

\begin{figure*}[h]
\centering
\small
\setlength{\tabcolsep}{3pt}
\renewcommand{\arraystretch}{1.08}

\begin{tabular}{r p{0.23\textwidth} p{0.55\textwidth} c}
\toprule
Step & State summary & Student action colored by teacher score
& Env. return \\
\midrule

1 &
searching for tomato &
\tokcell{take}{-0.10}
\tokcell{tomato}{-0.04}
\tokcell{2}{-0.08}
& 1 \\

2 &
holding tomato; needs cooling &
\tokcell{go}{0.20}
\tokcell{to}{0.14}
\tokcell{fridge}{0.32}
\tokcell{1}{0.18}
& 1 \\

3 &
holding cooled tomato; microwave needed &
\tokcell{go}{-0.07}
\tokcell{to}{-0.25}
\tokcell{microwave}{-1.70}
\tokcell{1}{-0.43}
& 1 \\

4 &
near microwave &
\tokcell{put}{-0.18}
\tokcell{tomato}{-0.11}
\tokcell{in}{-0.06}
\tokcell{microwave}{-0.15}
& 1 \\

\bottomrule
\end{tabular}

\vspace{2pt}
\caption{\textbf{Token-level teacher conflict on a successful ALFWorld trajectory.} Token scores indicate the teacher's preference for the student's generated actions. Despite receiving positive environment feedback, the student's correct transition toward the microwave receives strong negative teacher supervision.}
\label{fig:teacher-token-conflict}
\end{figure*}
\newtcolorbox{promptbox}[1]{
  enhanced,
  width=1\linewidth,
  sharp corners,
  boxrule=1pt,
  colframe=promptblue,
  colback=white,
  colbacktitle=promptblue,
  coltitle=white,
  title={#1},
  fonttitle=\bfseries\normalsize,
  fontupper=\normalsize,
  left=5mm,
  right=5mm,
  top=4mm,
  bottom=4mm,
  toptitle=1mm,
  bottomtitle=1mm,
  boxsep=0pt,
  before upper={
    \setlength{\parindent}{0pt}
    \setlength{\parskip}{0pt}
  }
}

\clearpage
\subsection{Prompt Templates}
\label{app:prompt}

We use a unified prompt format for agent interaction across environments.
The student receives task observations and action history, while the
teacher additionally accesses privileged skill information during OPD. The teacher receives the same student trajectory prefix, with privileged skill information prepended to the prompt.

\begin{figure}[H]
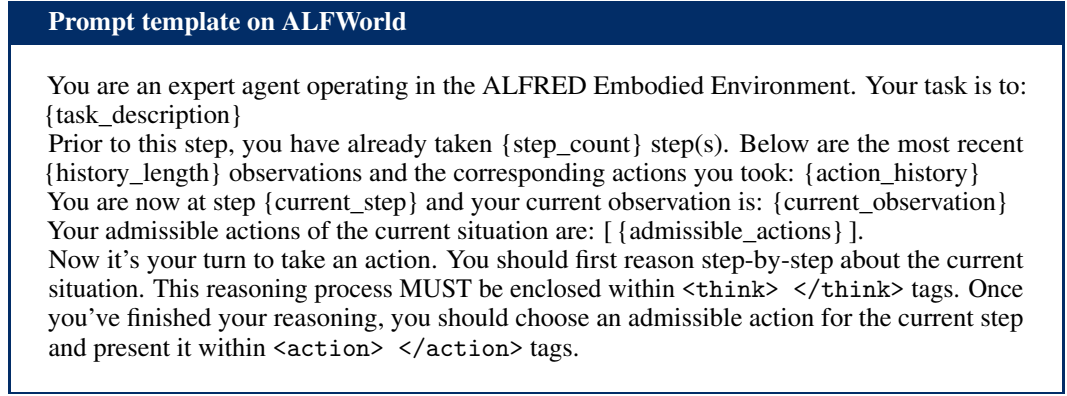

  \centering
  \begin{promptbox}{Prompt template on ALFWorld}
    You are an expert agent operating in the ALFRED Embodied
    Environment. Your task is to: \{task\_description\}\\
    Prior to this step, you have already taken \{step\_count\}
    step(s). Below are the most recent \{history\_length\}
    observations and the corresponding actions you took:
    \{action\_history\}\\
    You are now at step \{current\_step\} and your current
    observation is: \{current\_observation\}\\
    Your admissible actions of the current situation are:
    [\,\{admissible\_actions\}\,].\\
    Now it's your turn to take an action.
    You should first reason step-by-step about the current situation.
    This reasoning process MUST be enclosed within
    \texttt{\textless think\textgreater\
    \textless/think\textgreater} tags. Once you've finished your
    reasoning, you should choose an admissible action for the current
    step and present it within
    \texttt{\textless action\textgreater\
    \textless/action\textgreater} tags.
  \end{promptbox}
  \caption{Student prompt template for the ALFWorld task environment.}
  \vspace{5mm}
  \label{fig:student-alfworld-prompt}
\end{figure}

\begin{figure}[H]
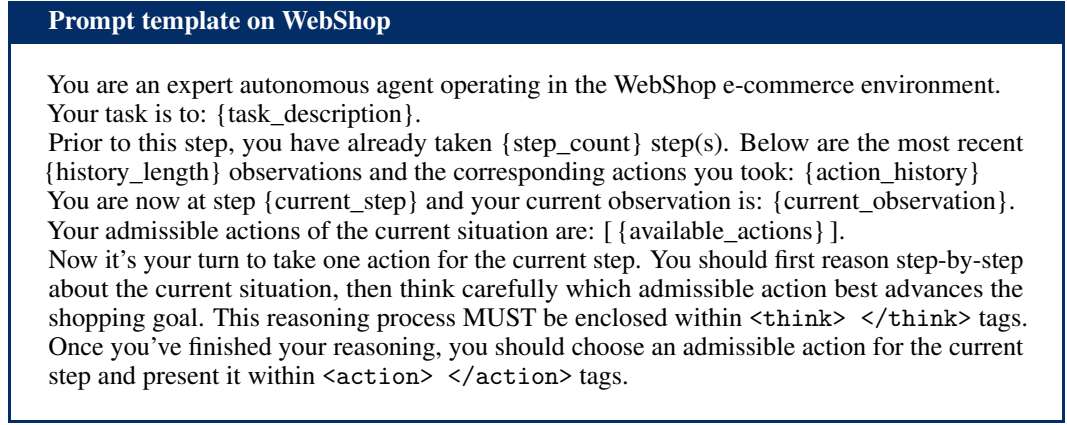

  \centering
  \begin{promptbox}{Prompt template on WebShop}
    You are an expert autonomous agent operating in the WebShop
    e-commerce environment.\\
    Your task is to: \{task\_description\}.\\
    Prior to this step, you have already taken
    \{step\_count\} step(s). Below are the most recent
    \{history\_length\} observations and the corresponding actions
    you took: \{action\_history\}\\
    You are now at step \{current\_step\} and your current
    observation is: \{current\_observation\}.\\
    Your admissible actions of the current situation are:
    [\,\{available\_actions\}\,].\\
    Now it's your turn to take one action for the current step.
    You should first reason step-by-step about the current situation,
    then think carefully which admissible action best advances the
    shopping goal. This reasoning process MUST be enclosed within
    \texttt{\textless think\textgreater\
    \textless/think\textgreater} tags. Once you've finished your
    reasoning, you should choose an admissible action for the current
    step and present it within
    \texttt{\textless action\textgreater\
    \textless/action\textgreater} tags.
  \end{promptbox}
  \caption{Student prompt template for the WebShop task environment.}
  \vspace{5mm}
  \label{fig:student-webshop-prompt}
\end{figure}

\begin{figure}[H]
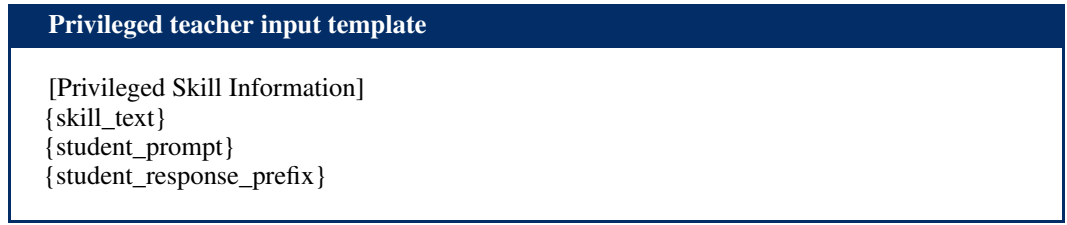

  \centering
  \begin{promptbox}{Privileged teacher input template}
    [Privileged Skill Information]\\
    \{skill\_text\}\\
    \{student\_prompt\}\\
    \{student\_response\_prefix\}
  \end{promptbox}
  \caption{Teacher-side input used for OPD.}
  \label{fig:teacher-privileged-input}
\end{figure}

\clearpage
\subsection{Training Dynamics}

\label{sec:training_dynamics}

\begin{figure}[H]

\begin{center}
\includegraphics[
      width=\linewidth,
      trim={0 0 0 30},
      clip
    ]{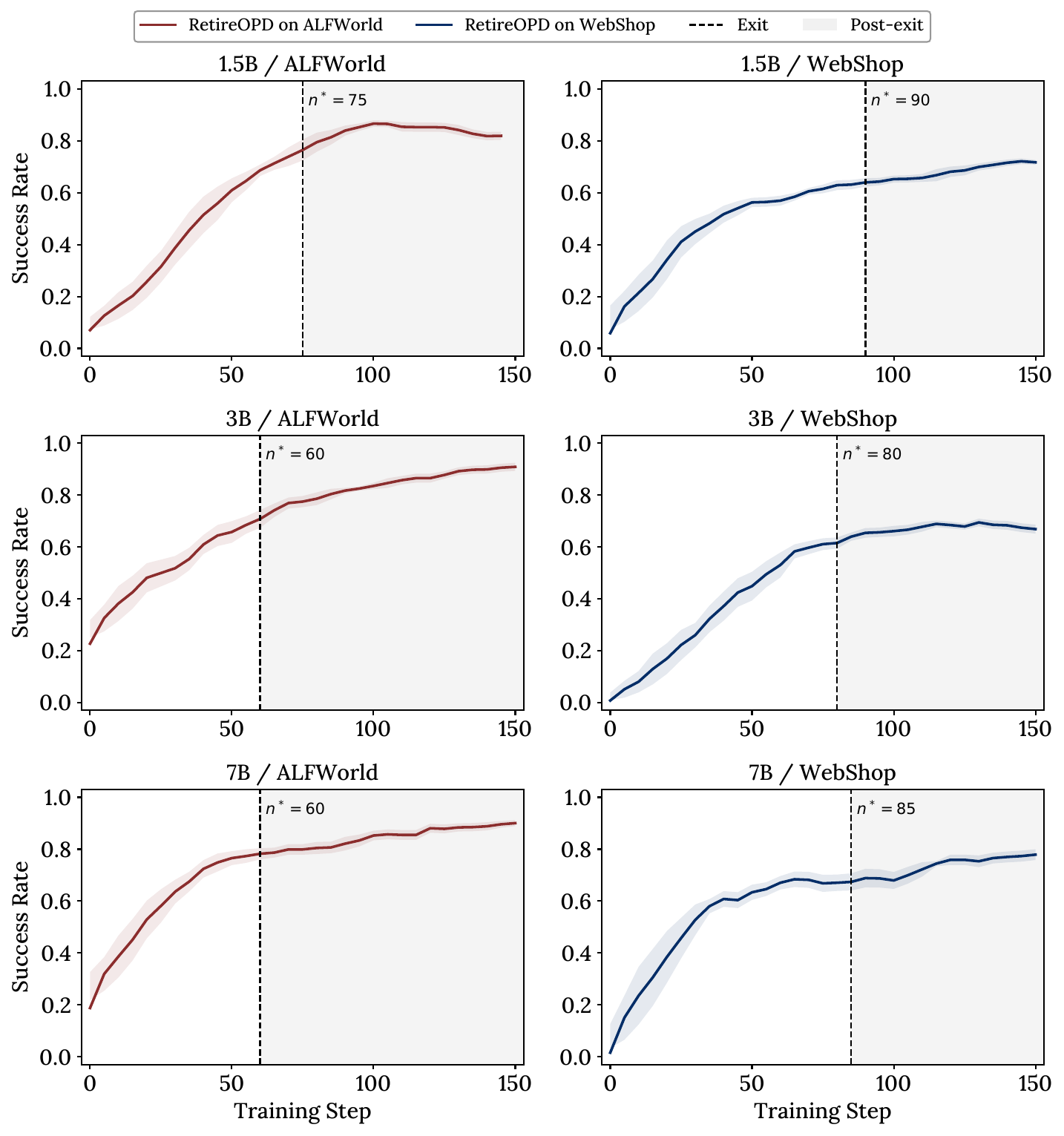}
\end{center}

\caption{\textbf{Validation Score} when training with Qwen2.5-1.5B-Instruct, Qwen2.5-3B-Instruct and Qwen2.5-7B-Instruct on ALFWorld and WebShop. \textit{n$^*$} denotes the retirement step.}
\label{fig:dynamics_success}
\end{figure}

\begin{figure}[t]

\begin{center}
\includegraphics[
      width=\linewidth,
      trim={0 0 0 30},
      clip
    ]{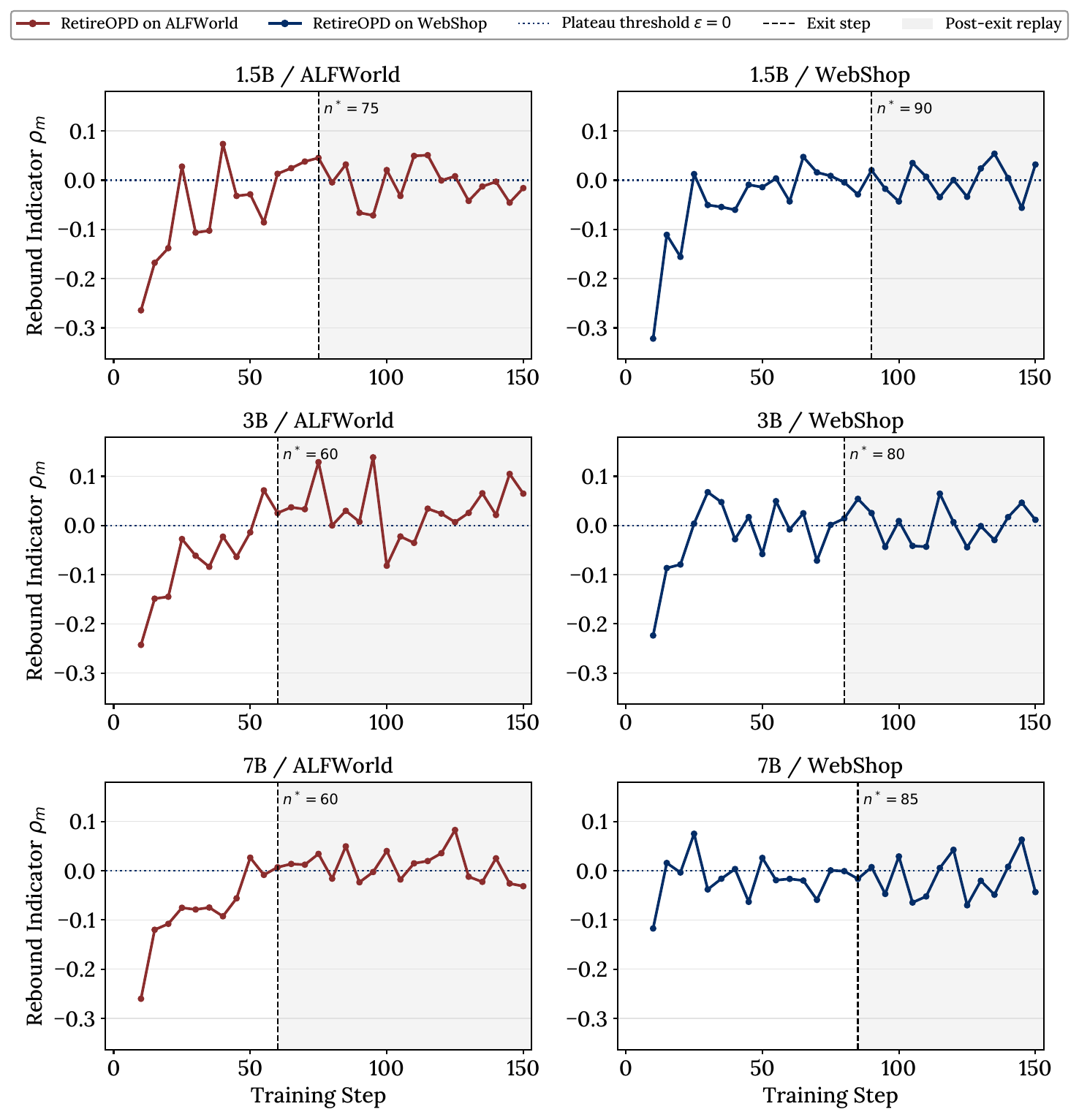}

\end{center}

\caption{\textbf{Rebound Indicator Curve} when training with Qwen2.5-1.5B-Instruct, Qwen2.5-3B-Instruct and Qwen2.5-7B-Instruct on ALFWorld and WebShop. \textit{n$^*$} denotes the retirement step, and Rebound Indicator threshold $\delta$ is 0.}
\label{fig:dynamics_discrepancy}
\end{figure}

\begin{figure}[t]

\begin{center}
\includegraphics[
      width=\linewidth,
      trim={0 0 0 30},
      clip
    ]{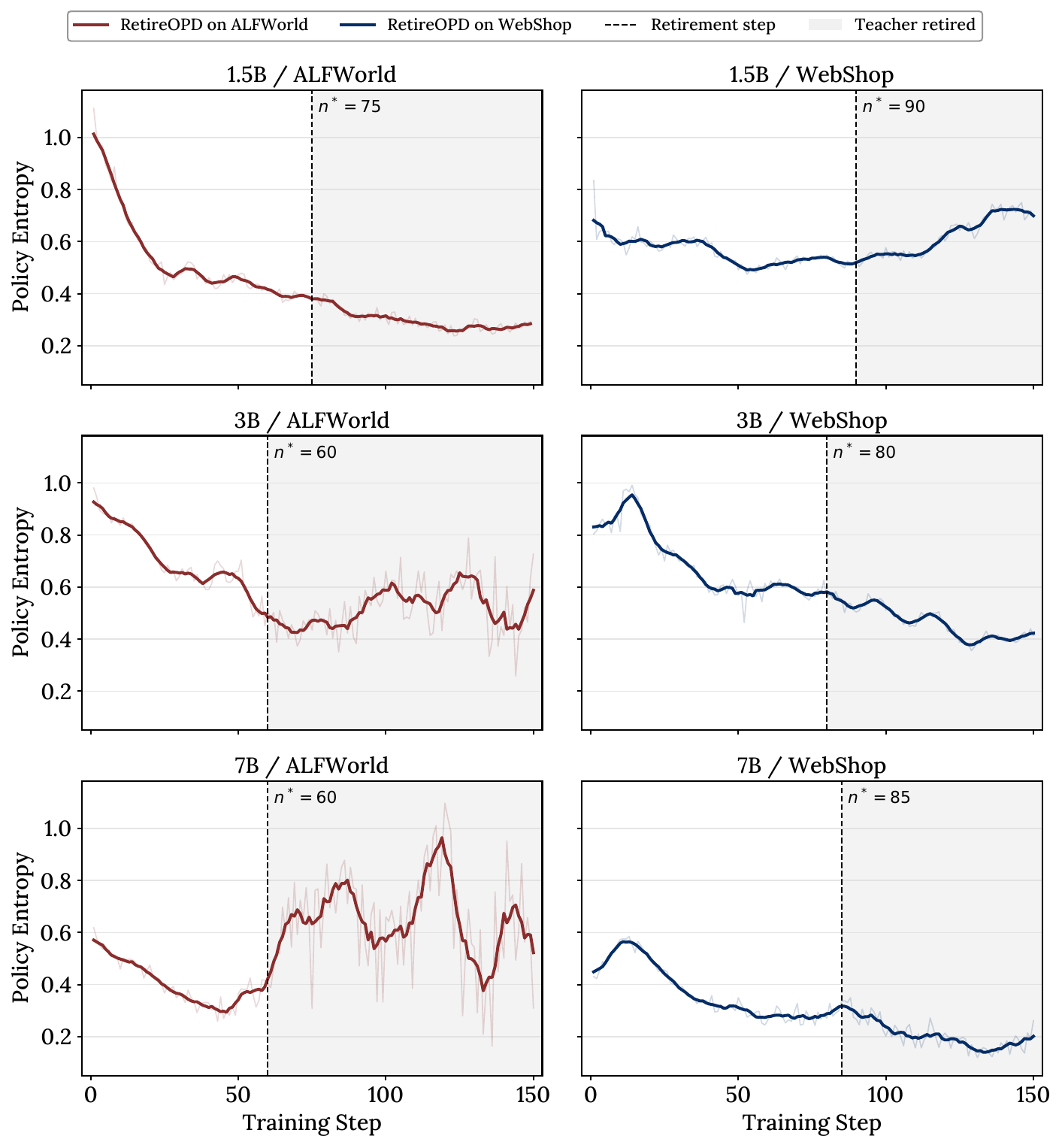}

\end{center}

\caption{\textbf{Entropy Curve} when training with Qwen2.5-1.5B-Instruct, Qwen2.5-3B-Instruct and Qwen2.5-7B-Instruct on ALFWorld and WebShop. \textit{n$^*$} denotes the retirement step.}
\label{fig:dynamics_entropy}
\end{figure}

\begin{figure}[t]

\begin{center}
\includegraphics[
      width=\linewidth,
      trim={0 0 0 30},
      clip
    ]{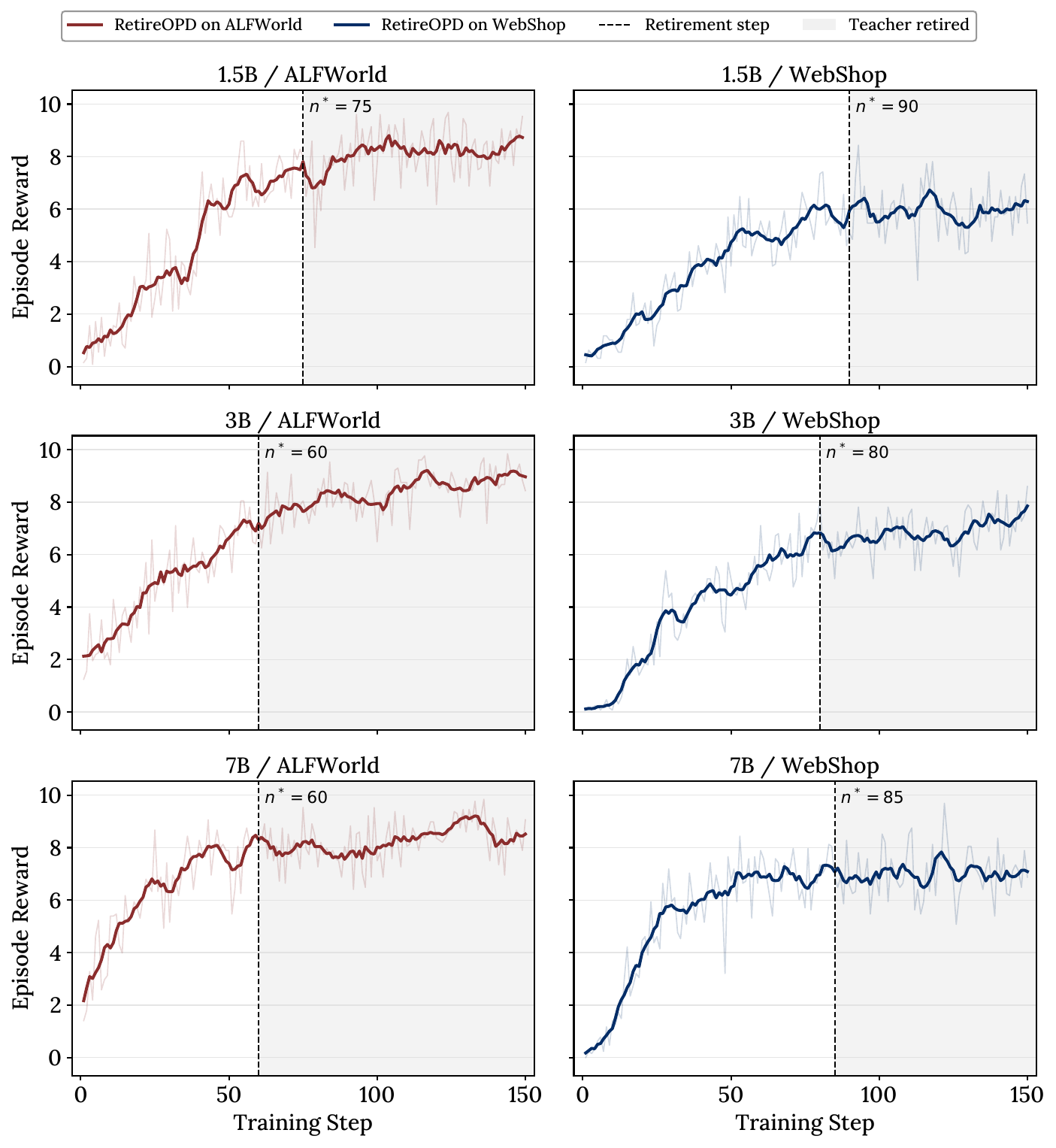}

\end{center}

\caption{\textbf{Reward Curve} when training with Qwen2.5-1.5B-Instruct, Qwen2.5-3B-Instruct and Qwen2.5-7B-Instruct on ALFWorld and WebShop. \textit{n$^*$} denotes the retirement step.}
\label{fig:dynamics_reward}
\end{figure}

We present the full training dynamics of \method{} across all model scales and environments in Figures~\ref{fig:dynamics_success}--\ref{fig:dynamics_reward}.

\newpage

\end{document}